\documentclass[10pt,journal,compsoc]{IEEEtran}
\ifCLASSOPTIONcompsoc
  \usepackage[nocompress]{cite}
\else
  \usepackage{cite}
\fi
\ifCLASSINFOpdf
\else
\fi

\usepackage{graphicx}
\usepackage{algorithmic}
\usepackage{adjustbox}
\usepackage{multirow}
\usepackage{mathtools}
\usepackage{bbm}
\usepackage{bm}
\usepackage{booktabs}
\usepackage{caption}
\usepackage{subcaption}
\usepackage{float}
\usepackage{changes}
\usepackage{amsmath}
\usepackage{amssymb}
\usepackage[ruled,vlined]{algorithm2e}
\usepackage{lineno}
\usepackage{pifont}
\usepackage{makecell}
\usepackage{color}
\usepackage{xcolor}

\newcommand{\bd}[1]{\textbf{#1}}
\def\DD{{\mathcal D}}

\newcommand{\R}[1]{#1}

\newcommand{\cmark}{{\ding{51}}}%
\newcommand{\xmark}{{\ding{55}}}%
\SetKwInput{KwInput}{Input}                
\SetKwInput{KwOutput}{Output}              
\SetKwInput{KwInit}{Initialization}         
\begin{document}
%
\title{Dynamic Loss For Robust Learning}
%
%
%
%

\author{Shenwang~Jiang{$^\dagger$}, 
        Jianan~Li{$^{\dagger,*}$}, 
        Jizhou Zhang,
	    Ying Wang,
	    and~Tingfa~Xu{$^*$}
\thanks{This work was financially supported by the National Natural Science Foundation of China (No. 62101032), the Postdoctoral Science Foundation of China (Nos. 2021M690015, 2022T150050), and Beijing Institute of Technology Research Fund Program for Young Scholars (No. 3040011182111).}
\thanks{S. Jiang, J. Li, J. Zhang , Y. Wang, and T. Xu are with Beijing Institute of Technology, Beijing 100081, China. E-mail: jiangwenj02@gmail.com, \{ lijianan, zhangjizhou, 3120215325,  ciom\_xtf1\}@bit.edu.cn. T. Xu is also with Big Data and Artificial Intelligence Laboratory, Beijing Institute of Technology Chongqing Innovation Center, Chongqing 401135, China.}
\thanks{$^\dagger$ Shenwang Jiang and Jianan Li contributes equally to this work.}
\thanks{$^{*}$ Jianan Li and Tingfa~Xu are co-corresponding authors.}
\thanks{The code will present in https://github.com/jiangwenj02/dynamic\_loss.}
}

%
%

\markboth{Journal of \LaTeX\ Class Files,~Vol.~14, No.~8, August~2015}%
{Shell \MakeLowercase{\textit{et al.}}: Bare Demo of IEEEtran.cls for Computer Society Journals}
%



\IEEEtitleabstractindextext{%
\begin{abstract}
Label noise and class imbalance are common challenges encountered in real-world datasets. Existing approaches for robust learning often focus on addressing either label noise or class imbalance individually, resulting in suboptimal performance when both biases are present.
To bridge this gap, this work introduces a novel meta-learning-based dynamic loss that adapts the objective functions during the training process to effectively learn a classifier from long-tailed noisy data.
Specifically, our dynamic loss consists of two components: a label corrector and a margin generator. The label corrector is responsible for correcting noisy labels, while the margin generator generates per-class classification margins by capturing the underlying data distribution and the learning state of the classifier.
In addition, we employ a hierarchical sampling strategy that enriches a small amount of unbiased metadata with diverse and challenging samples. This enables the joint optimization of the two components in the dynamic loss through meta-learning, allowing the classifier to effectively adapt to clean and balanced test data.
Extensive experiments conducted on multiple real-world and synthetic datasets with various types of data biases, including CIFAR-10/100, Animal-10N, ImageNet-LT, and Webvision, demonstrate that our method achieves state-of-the-art accuracy.  
\end{abstract}

}

\maketitle

\IEEEdisplaynontitleabstractindextext

%
\IEEEpeerreviewmaketitle

\IEEEraisesectionheading{\section{Introduction}\label{sec:introduction}}

%
%
%
%

\IEEEPARstart{D}{eep} {neural networks (DNNs) have demonstrated remarkable success attributed to the abundance of labeled data \cite{simonyan2014very, he2016deep, ding2021repvgg}. However, real-world datasets often exhibit long-tailed distributions and inevitably contain noisy labels \cite{li2017webvision}. The presence of such biased data distributions renders DNNs susceptible to overlooking tail classes \cite{cao2019learning} and memorizing noisy training labels \cite{jiang2018mentornet}, leading to suboptimal performance on balanced and clean test data. Consequently, addressing the challenge of robust learning from long-tailed data with noisy labels has received growing attention in recent studies \cite{jiang2021delving, cao2020heteroskedastic}.}

\begin{figure}[t]
 \centering
  \begin{subfigure}[b]{0.95\linewidth}
  \includegraphics[width=\linewidth]{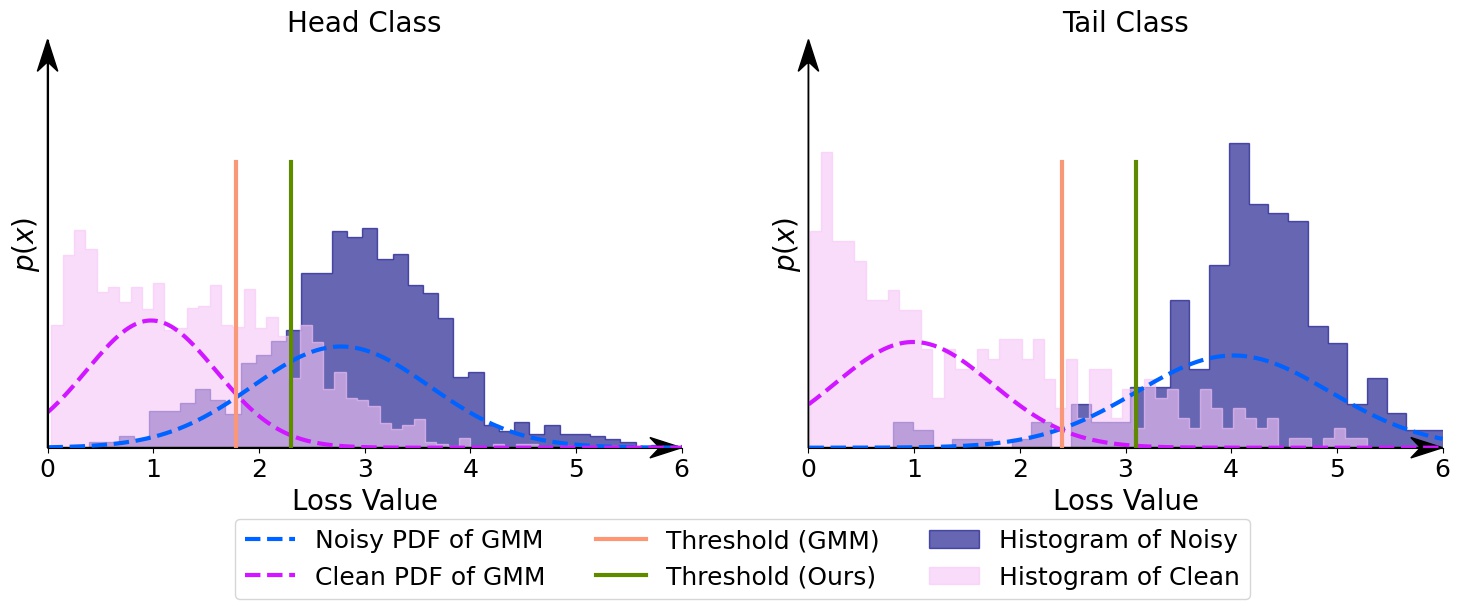}
  \end{subfigure} \\
  \begin{subfigure}[b]{0.95\linewidth}
   \includegraphics[width=\linewidth]{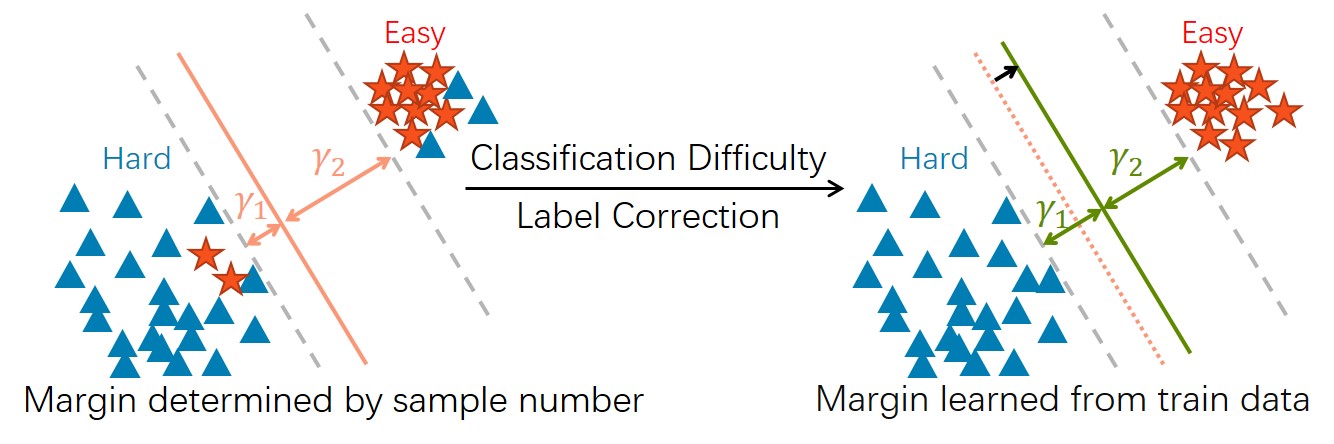}
  \end{subfigure}
  \vspace{-0.35cm}
  \caption{
  {\bd{Upper:} Conventional class-specific Gaussian Mixture Model (GMM) with fixed hyperparameters tends to misclassify clean samples into the noisy split, whereas our label corrector learns optimal division thresholds for different classes using training data and samples' class-specific loss rank.
  \bd{Lower}: Previous methods generate classification margin ($\gamma$) based solely on sample number, whereas our margin generator takes into account the presence of noisy labels and the distinct classification difficulties of different classes, resulting in a more appropriate margin.}
  }
    \label{fig:noise_margin}
\vspace{-0.55cm}
 \end{figure}

{Recent advancements in robust learning \cite{shu2019meta, jiang2018mentornet, kang2019decoupling, liu2019large} have explored various techniques, including correcting noisy labels \cite{zheng2021meta, li2020dividemix} and adjusting classification margins \cite{cao2019learning, ren2020balanced}, to address the issues of label noise and class imbalance, respectively. However, these solutions heavily rely on manually-designed rules to distinguish between noisy and clean samples, or pre-set parameters to obtain prior knowledge of class distribution. When label noise meets class imbalance, manual interventions become impractical since tail classes make it difficult to identify label noise, while noisy labels make observed class distribution unreliable. As a result, we aim to pave a new path for robust learning from biased data in a fully self-adaptive manner.}

{Taking the aforementioned challenges into consideration, we propose a novel meta-learning-based dynamic loss that comprises of a learnable \textit{label corrector} and a \textit{margin generator}. Our approach aims to learn a classifier robustly from long-tailed noisy data. In contrast to the predefined fixed objective functions adopted by priors, our dynamic loss learns to correct per-sample labels and adjust per-class classification margins simultaneously by perceiving the underlying data distribution and the learning state of the classifier. As a result, our approach provides suitable dynamic learning objectives throughout the training process.}

{The main challenge in correcting noisy labels lies in identifying noisy samples and disregarding or reusing them appropriately. Previous solutions \cite{li2020dividemix, xu2021faster} have used a GMM with fixed hyperparameters to differentiate between noisy and clean samples based on their loss values, which is inadequate for long-tailed noisy data. This is because some clean samples of tail classes may have higher loss values than noisy samples of head classes due to the highly skewed class distribution. While a class-specific GMM \cite{Huang_Bai_Zhao_Bai_Wang_2022} is a straightforward solution, it is still prone to confusion due to the significant variation in noise rates among different classes (Figure~\ref{fig:noise_margin}).
Furthermore, previous methods have relabeled identified noisy samples by combining their assigned labels and the classifier's predictions in a fixed manner. This approach can lead to incorrect corrections since the classifier is imprecise at the beginning of training and susceptible to overfitting to noisy samples towards the end. To address these issues, we propose a novel label corrector that can learn to jointly identify and relabel noisy samples in a fully learnable fashion. This is achieved by taking into account both the class-specific loss rank of samples and the learning state of the classifier.}

{In order to adjust per-class classification margins, we are inspired by the fact that classes with fewer samples are associated with larger generalization error bounds, which can be minimized by increasing the classification margin~\cite{cao2019learning}. However, most prior approaches \cite{ren2020balanced} pre-define a fixed margin for each class based solely on its sample number (Figure~\ref{fig:noise_margin}). This approach suffers from two main drawbacks: i) the per-class sample number becomes unreliable in the presence of noisy labels; and ii) the distinct classification difficulties among different classes are simply ignored. To address these limitations, we propose a novel margin generator that can learn to produce a suitable dynamic margin for each class by self-perceiving the true class distribution underlying the noisy data, as well as the classification difficulty of each class.}

{To enable the learning of the classifier from long-tailed noisy data, we propose a unified dynamic loss that integrates the label corrector and the margin generator, and optimize them through meta-learning. This approach allows the objective function for the classifier to be dynamically adjusted throughout the training process. The label corrector and the margin generator work in tandem to improve the learning from noisy data. Specifically, the label corrector restores the true distribution of the data, which allows the margin generator to produce a more suitable classification margin. In turn, the resulting improved margin boosts the accuracy of the predicted labels and enhances the reliability of the label corrector.}
{Furthermore, the convergence of meta-learning is widely recognized as a challenging issue in the field~\cite{antoniou2019train}. In order to tackle this problem, we propose the incorporation of a group optimization strategy and the explicit utilization of known information. These techniques serve to streamline the meta-learning task, mitigate input instability, and ultimately enhance the convergence of the meta-learning process.}

{To collect a small amount of unbiased meta data for meta-learning, we have developed a new hierarchical sampling strategy that progressively builds a random primary set and then a balanced clean meta set. This approach ensures that the meta set is enriched with diverse and challenging samples that better simulate the distribution of real test data, thereby circumventing the problem of over-reliance on straightforward samples. By doing so, our dynamic loss can guide the classifier learning robustly on various types of biased data in a fully self-adaptive manner.}

{We conduct a comprehensive evaluation of our proposed dynamic loss on both synthetic and real-world long-tailed data with label noise, achieving state-of-the-art results on a wide variety of benchmarks featuring various imbalance ratios and noise rates, such as CIFAR-10/100~\cite{krizhevsky2009learning}, Animal-10N~\cite{song2019selfie}, ImageNet-LT~\cite{liu2019large}, and Webvision~\cite{li2017webvision}. Furthermore, additional tests conducted on purely imbalanced or noisy data further validate the dynamic loss's exceptional adaptability and robustness.}

{In summary, our work contributes in the following ways:
\begin{itemize}
\item We introduce a straightforward yet powerful dynamic loss, which paves a new way for robust learning on various forms of biased data in a fully self-adaptive manner. 
\item We devise a novel hierarchical sampling strategy that efficiently generates diverse and unbiased meta data, which allows for better simulation of the true data distribution and improves the generalization ability of the model.
\item We achieve state-of-the-art performance on multiple synthetic and real-world datasets, demonstrating the effectiveness and versatility of our proposed approach.
\end{itemize}}

\section{Related Works}

\noindent\textbf{Long-Tailed Learning.}
{Previous works on long-tailed learning can be broadly categorized into three main approaches: data re-sampling~\cite{chawla2002smote,buda2018systematic}, boundary adjustment~\cite{tang2020long}, and re-weighting~\cite{shu2019meta,lin2017focal,huang2016learning}. The data re-sampling approach involves balancing the class distribution by over-sampling the tail classes. However, this method is prone to overfitting on the tail classes. Methods belonging to the second category aim to enlarge the classification boundary of the tail classes while narrowing that of the head classes. This is achieved by modifying the classification threshold~\cite{menon2020long} or by adjusting the weights of the output layer through normalization~\cite{kang2019decoupling}. The re-weighting approach aims to assign larger loss weights to the tail classes. Conventional approaches of this category~\cite{huang2016learning,huang2019deep} impose weights on each training sample directly, which may cause unstable training due to sensitivity to outliers~\cite{ren2020balanced}. Recent works~\cite{tan2020equalization} modify the predicted scores in the Softmax function to achieve re-weighting, which yields more stable training and promising performance. In this work, we adapt the re-weighting strategy to more challenging long-tailed scenarios with label noise.}

\noindent\textbf{Learning under Label Noise.}
{There are two main categories of methods for learning under label noise: sample re-weighting and relabeling. The re-weighting strategy involves assigning lower weights to samples with larger loss values, which are considered to be noisy~\cite{kumar2010self,huang2019o2u}. MentorNet~\cite{jiang2018mentornet} learns data-driven curriculums for deep convolutional neural networks trained on corrupted labels. Meta-Weight-Net~\cite{shu2019meta} learns an explicit weighting function directly from a small set of clean data.  MetaSeg~\cite{jiang2023metaseg} directly generates weights according to the feature  and given label of images. On the other hand, the relabeling strategy leverages noisy samples by refining their labels. Bootstrapping~\cite{reed2014training} integrates assigned labels and model predictions through interpolation. Some works divide clean and noisy samples based on priors learned from a manually generated noisy set~\cite{chen2021sample} and then take advantage of noisy samples~\cite{berthelot2019mixmatch}.}

\noindent\textbf{Long-tailed Learning under Label Noise.}
{Recently, there have been several approaches proposed for addressing long-tailed learning with label noise. For instance, HAR~\cite{cao2020heteroskedastic} applies a data-dependent regularization technique to regularize different regions of the input space differently. CurveNet~\cite{jiang2021delving} learns to assign appropriate weights to different samples based on their loss curves. ROLT~\cite{wei2021robust} combines DivideMix and LDAM to correct noisy labels and improve the performance of tail classes. However, unlike these methods, the approach presented in this work is the first to simultaneously correct noisy labels and adaptively adjust per-class classification margins in a learnable and adaptive manner according to the training data.}

\begin{figure*}[t]
\centering
   \includegraphics[width=0.95\linewidth]{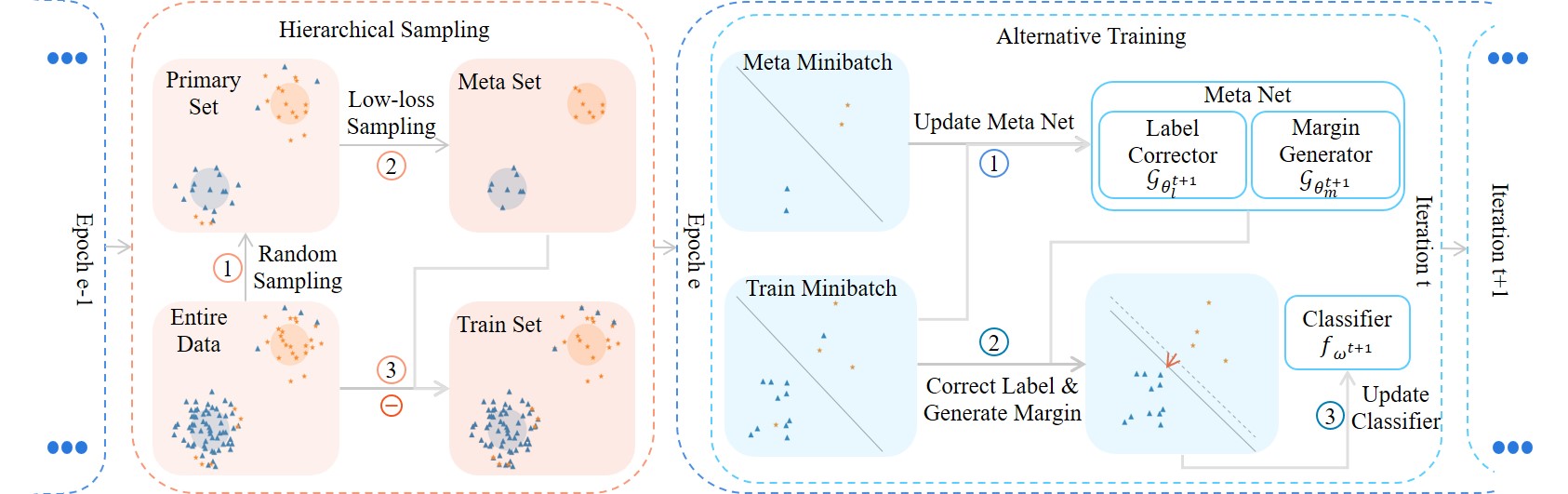}
  \caption{{Overview of our learning paradigm using the dynamic loss. Each training epoch involves splitting the entire data into an unbiased meta set and a biased training set. In each iteration, we jointly update the label corrector and margin generator using meta-learning on mini-batch meta and training data, followed by updating the classifier on mini-batch training data by minimizing the dynamic loss using corrected per-sample labels and generated per-class margins.}}
\label{fig:overview}
\end{figure*}


\section{Methods}
{In this section, we will provide a detailed description of our dynamic loss and the process of optimizing it through meta-learning.}

\subsection{Overview}
{Our objective is to train a classifier $\bm{f}_{\bm{\omega}}$ with learnable parameters $\bm{\omega}$ using a noisy and imbalanced training set $\mathcal{D}=\{\bm{x}_i,\bm{y}_i\}_{i=1}^N$, where each training example comprises an image $\bm{x}_i$ and its corresponding one-hot class label $\bm{y}_i$. Although the training set is beset by label noise and class imbalance, our aim is to ensure that the classifier accurately recognizes all classes. To achieve this goal, we use a balanced and clean test set.}

{We optimize the learnable parameters $\bm{\omega}$ by minimizing the classification loss on the training set, given by
\begin{equation}
    \bm{\omega}^* = \mathop{\arg\min}_{\bm{\omega}} \sum_{i=1}^N \ell (\bm{y}_i, \  \bm{f}_{\bm{\omega}}(\bm{x}_i)).
    \label{eq:ce}
\end{equation}
where $\ell(\cdot)$ represents the cross-entropy loss. However, due to the presence of label noise and class imbalance in the training set, optimizing using this naive cross-entropy loss suffers from two significant drawbacks. Firstly, the labels assigned to noisy samples do not correspond to their ground-truths, leading to high loss values and causing the model to memorize the noisy labels. Secondly, tail classes occur much less frequently than head classes, but have the same classification margins, making them susceptible to poor generalization.}

{To address the aforementioned issues, we introduce a novel dynamic loss function that corrects noisy labels and adjusts the classification margins for different classes in an adaptive and learnable manner. The dynamic loss is given by:
\begin{equation}
    \mathcal{DL} = \ell(\bm{y}^*_i, \  \bm{f}_{\bm{\omega}}(\bm{x}_i) + \bm{q}),
    \label{eq:dynamic}
\end{equation}
where $\bm{y}^*_i$ and $\bm{q} \in \mathbb{R}^C$ represent the reassigned label and the additive classification margin for $\bm{x}_i$, respectively.}

{As depicted in Figure~\ref{fig:overview}, the dynamic loss incorporates a learnable label corrector $\mathcal{G}_{{\bm{\theta}}_l}$, parameterized by ${\bm{\theta}}_l$, and a margin generator $\mathcal{G}_{{\bm{\theta}}_m}$, parameterized by ${\bm{\theta}}_m$, which are responsible for correcting per-sample labels and per-class classification margins, respectively. We jointly optimize these components along with the classifier $\bm{f}_{\bm{\omega}}$ using meta-learning.
In the following, we elaborate on the two components and their optimization in detail.}

\subsection{Label Corrector} \label{sec:noisy}

{The label corrector operates in a class-wise manner to detect and correct wrongly assigned labels in noisy samples. To identify such samples, it first divides all samples into $C$ groups based on their class and sorts the samples in each group individually according to their loss values. It then divides the sorted samples in each group into $R$ bins of equal size. To determine whether a bin $r \in \{1, \dots, R\}$ is dominated by noisy or clean samples, the label corrector uses a lightweight class-wise meta network. As a result, the bin index $r_i$ of the loss value for sample $i$ can be used as a reliable indicator for identifying label noise.}

{In terms of label correction, the classifier can learn from the clean samples that dominate the data and transfer this knowledge to the noisy samples, provided that the classifier has not severely over-fitted on biased data. This allows the classifier's predictions on noisy samples to be more accurate and closer to their ground-truth labels, facilitating the correction of wrongly assigned labels.}

{Drawing from the aforementioned observation, we present our label corrector which reassigns a ground-truth label $\bm{y}^*_i$ for a given sample $\bm{x}_i$ by calculating a weighted sum of its assigned label $\bm{y}_i$ and the prediction $\bm{y}'_i$ made by the classifier. The weights used for the summation are dependent on the loss bin index $r_i$, given by:
\begin{equation}
\begin{split}
    \bm{y}^*_i & = \mathcal{G}_{{\bm{\theta}}_l}(\bm{y}_i,\bm{y}'_i,r_i) \\ 
     & = \bm{y}_i*g(r_i | y_i) + \bm{y}'_i*(1-g(r_i |  y_i)),
    \label{eq:label_correction}
\end{split}
\end{equation}
The function $g:r_i | y_i \to [0,1]$ is a class-dependent weighting function that maps the bin index $r_i$ to a balance weight and is learned by a small meta network. The network is comprised of a one-hot encoder and a two-layer perceptron (MLP) with a Sigmoid activation function.}

{In instances where sample $i$ is regarded as noisy due to a high loss value resulting in a large bin index $r_i$, the computed value of $g(r_i |  y_i)$ tends to approach zero. Consequently, the label corrector adjusts its label assignment by incorporating the prediction made by the classifier to rectify its initial erroneous label assignment.
The reverse is also true: if the sample has a low loss value and a small bin index, the weight assigned to the classifier's prediction is close to zero, allowing the assigned label to remain dominant in the label correction process.}

\subsection{Margin Generator} \label{sec:method:imb}
{To design the margin generator, we revisit the Label-Distribution-Aware Margin Loss (LDAM)~\cite{cao2019learning} from the perspective of generalization error bound. Given that tail classes often have fewer training samples, they typically exhibit larger generalization error bounds when compared with head classes. Since the generalization error bound is often negatively correlated with the magnitude of the classification margin, increasing the classification margins for the tail classes can effectively reduce their generalization error bounds.}

{In this regard, Balanced Meta-Softmax~\cite{ren2020balanced} presents an unbiased extension of standard Softmax by adjusting the classification margin for class $j$ based on its sample number $n_j$, and adding the margin $-log(n_j)$ to the confidence score $p_j$ predicted by the classifier:
\begin{equation}
    {\ell}'(\omega|y=j) =-log(\frac{e^{p_j + log(n_j)}}{\sum_{i=1}^C e^{p_i + log(n_i)}}).
    \label{eq:metasoftmax}
\end{equation}}

{However, when dealing with long-tailed data with noisy labels, the sample number $n_j$ may not accurately reflect the true number of samples belonging to class $j$ due to the existence of label noise. Furthermore, manually defining the margin solely based on the sample number may not account for the distinct classification difficulties among different classes.}

{We propose a learnable margin generator $\mathcal{G}_{{\bm{\theta}}_m}$, which consists of a two-layer multi-layer perceptron (MLP) that can dynamically adjust the margin for each class. During classifier training, the margin generator optimizes a learnable margin vector $\bm q \in \mathbb{R}^C$, which is initialized with an all-ones vector:
\begin{equation}
    \bm{q}=\mathcal{G}_{{\bm{\theta}}_m}(\bm{1})=[q_1, \dots, q_C].
    \label{eq:margin}
\end{equation}
By integrating the margin vector into the standard Softmax loss, we have the modified loss function:
\begin{equation}
    \hat{\ell}(\omega|y=j)=-log(\frac{e^{p_j + q_j}}{\sum_{i=1}^C e^{p_i + q_i}}).
    \label{eq:souroftmax}
\end{equation}
where $p_j$ and $q_j$ denote the predicted score and the learned margin for class $j$, respectively.
Since the classification margin is $-{q_j}$ in our formulation, the learned margin $q_j$ for class $j$ should be positively correlated with its sample number.}

{Hence the margin generator can automatically adjust per-class margins by adapting to the underlying true class distribution in long-tailed noisy data, as well as the classification difficulty of each class. Importantly, the margin generator can accomplish this in a learnable manner, without requiring any manual intervention or prior information.}

\subsection{Hierarchical Sampling Strategy}
{We combine the label corrector $\mathcal{G}_{{\bm{\theta}}_l}$ and margin generator $\mathcal{G}_{{\bm{\theta}}_m}$ into a unified meta net $\mathcal{G_{\bm{\theta}}}$, which is a crucial element of our dynamic loss. We employ meta-learning to optimize $\mathcal{G_{\bm{\theta}}}$ and facilitate the learning of the classifier $\bm{f}_{\bm{\omega}}$ to better adjust to balanced and clean test data.}

{To enable meta-learning, a meta set $\mathcal{D}_c=\{(\bm{x}_i,\bm{y}_i)\}_{i=1}^{M}$ containing a small number of balanced and clean data must be constructed.
A simple approach is to select $M_1$ samples with the lowest classification loss values from each class in $\mathcal{D}$, as samples with lower loss values computed by $\bm{f}_{\bm{\omega}}$ are more likely to have correctly assigned labels.
However, this approach can be problematic because easier samples often have lower loss values during training, which can result in fixed easy samples being selected at each epoch and increase the risk of overfitting to such samples.}

{Therefore, we have devised a hierarchical sampling strategy to create $\mathcal{D}_c$. This strategy involves a two-step process: first, we randomly select $M_0$ samples from each class in $\mathcal{D}$ to construct a \emph{primary set}; second, we choose $M_1$ samples with low loss from each class in the primary set to form the final \emph{meta set}. The samples that are not selected for the meta set are included in the counterpart set $\DD_n$. Refer to Fig.~\ref{fig:overview} for an illustration of this process.}

{The advantages of incorporating the primary set are two-fold. Firstly, since the samples in the primary set are randomly selected at each epoch, it ensures that the resulting meta set is distinct across different epochs. Secondly, the primary set contains fewer samples than $\mathcal{D}$, increasing the likelihood of selecting hard yet clean samples located near the decision boundary into the meta set. This hierarchical sampling strategy guarantees the dynamism and diversity of the meta set, preventing the model from overfitting to biased data.}

\begin{figure}[t]
\centering
   \includegraphics[width=1.0\linewidth]{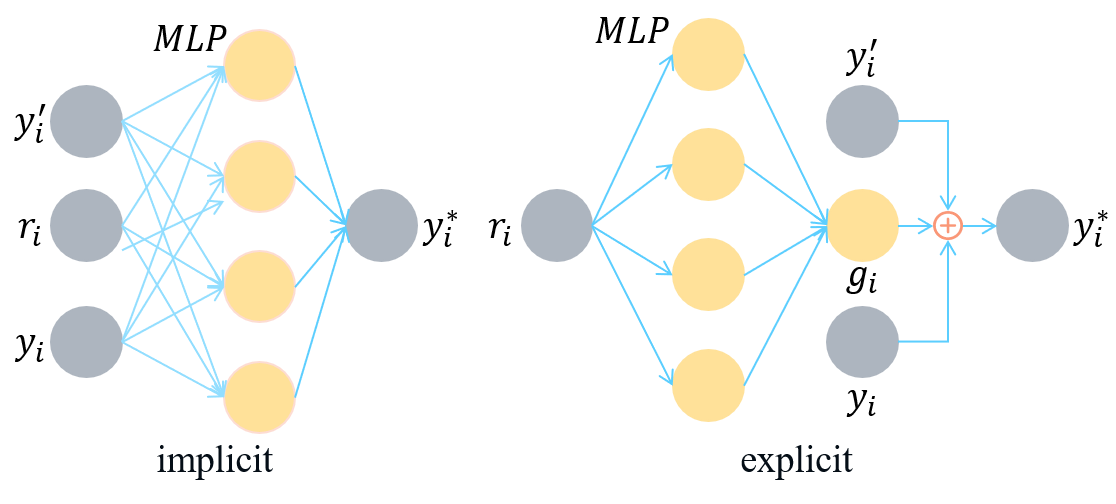}
   \caption{{Visualization of implicit and explicit inputs. $\bm{y}_i$, $\bm{y}'_i$, $\bm{y}^*_i$ and $r_i$ represent the assigned label of the training set, predicted label of the classifier, reassigned label of our meta net, and the bin index of the loss rank, respectively.}}
\label{fig:vis:implicit}
\end{figure}

\begin{table*}[t]
  \centering
  \setlength{\tabcolsep}{12.0pt}
  \renewcommand\arraystretch{1.2}
  \caption{{Detailed settings for training on different datasets.}}
  \label{tab:training_details}
    \begin{tabular}{c|c|c|c|c|c|c}
    \toprule
    \multicolumn{2}{c|}{\textbf{Settings}} & \textbf{CIFAR-10} & \textbf{CIFAR-100} & \textbf{Webvision} & \textbf{Animal-10N} & \textbf{ImageNet-LT} \\
    \midrule
    \multirow{5}{*}{\textbf{Classfier}} & Optimizer & \multicolumn{5}{c}{SGD} \\
    \cline{2-7}          & Momentum & \multicolumn{5}{c}{0.9} \\
    \cline{2-7}          & Weight Decay & 5e-4 & 5e-4 & 1e-4 & 5e-4 & 1e-4 \\
    \cline{2-7}          & Learning Rate & 0.1   & 0.1   & 0.02  & 0.02  & 0.1 \\
    \cline{2-7}          & Learning Scheduler & \multicolumn{5}{c}{Cosine Annealing} \\
    \midrule
    \multirow{4}{*}{\textbf{MetaNet}} & Optimizer & \multicolumn{5}{c}{Adam} \\
    \cline{2-7}          & Weight Decay & \multicolumn{5}{c}{0} \\
    \cline{2-7}          & Learning Rate & \multicolumn{5}{c}{3e-3} \\
    \cline{2-7}          & Learning Scheduler & \multicolumn{5}{c}{Fixed} \\
    \midrule
    \multirow{6}{*}{\textbf{Others}} & M0    & 0.5   & 0.5   & 0.5   & 0.5   & - \\
    \cline{2-7}          & M1    & 0.25  & 0.25  & 0.25  & 0.25  & - \\
    \cline{2-7}          & Epoch & 300   & 300   & 150   & 100   & 90 / 400 \\
    \cline{2-7}          & Warmup Epoch & 5     & 5     & 1     & 5     & 0 \\
    \cline{2-7}          & Batch Size & 512   & 512   & 64    & 128   & 128 \\
    \cline{2-7}          & Rank bins & 100   & 50    & 100   & 100   & - \\
    \bottomrule
    \end{tabular}%
\end{table*}%

\subsection{\R{Optimization}}

{The convergence of meta-learning poses significant challenges~\cite{antoniou2019train}. Previous approaches~\cite{shu2019meta,xu2021faster,jiang2021delving} employ a strategy of inputting all known information, particularly given labels, predicted labels and loss value, into the meta net, generating customized weights for each sample. However, due to computational complexity, meta nets often exhibit a relatively simple network structure consisting of a few fully connected layers, limiting their capability to handle complex tasks. Notably, in scenarios involving biased data, a substantial portion of the known information, such as predicted and given labels, can be omitted from the meta net's input. This is because only one of these labels needs to be selected as the sample's label, and a weight can be learned to combine the two labels and obtain a new label. Directly incorporating these known information into the meta net not only increases the optimization difficulty but also diminishes the utilization of the information. This is due to the implicit nature of using these information, requiring the meta net to extract it from the input, as expressed by:
\begin{equation}
    \bm{y}^*_i = \mathcal{G}_{{\bm{\theta}}_l}(\bm{y}_i,\bm{y}'_i,r_i) = g(\bm{y}_i,\bm{y}'_i,r_i).
    \label{eq:imp}
\end{equation}}

{To enhance the utilization of known information, we opt for explicit utilization through our Label Corrector, as illustrated in Figure~\ref{fig:vis:implicit} and governed by Equation~\ref{eq:label_correction}.}

{The customization of weights for individual samples diminishes the impact of diverse input information on the meta net. However, input information often encompasses errors, such as varying sample loss values during training and overlapping loss values between noisy and clean samples. These factors make it challenging to discern between them solely based on loss values. To overcome this challenge, we employ a grouping strategy that involves sorting by loss value for noisy data and given labels for imbalanced data. This strategy brings stability to the input information, reducing its instability and enhancing the influence of each input on the meta net. By explicitly utilizing known information and implementing the grouping strategy, we simplify the task of the meta net, promoting easier convergence and the assignment of accurate weights and margins to each sample group.}

\begin{table*}[h]
  \centering
  \setlength{\tabcolsep}{6.0pt}
  \renewcommand\arraystretch{1.1}
  \caption{{Accuracy (\%) on CIFAR-10-N-LT with varying imbalance ratio and noise rate.}}
  \label{tab:sota-cifar10}%
  \resizebox{\textwidth}{!}{
    \begin{tabular}{lc|ccccc|ccccc|c}
        \toprule
    \multicolumn{2}{c|}{\textbf{Imbalance Ratio}} & \multicolumn{5}{c|}{\textbf{10}}               & \multicolumn{5}{c|}{\textbf{100}}              & \multirow{2}[4]{*}{\textbf{Avg.}} \\
\cmidrule{1-12}    \multicolumn{2}{c|}{\textbf{Noise Rate}} & \textbf{0.1}   & \textbf{0.2}   & \textbf{0.3}   & \textbf{0.4}   & \textbf{0.5}   & \textbf{0.1}   & \textbf{0.2}   & \textbf{0.3}   & \textbf{0.4}   & \textbf{0.5}   &  \\
    \midrule
    \multirow{2}[2]{*}{Cross Entopy} & Last  & 76.63  & 68.89  & 61.21  & 54.01  & 44.09  & 59.99  & 50.56  & 44.81  & 37.13  & 30.02  & 52.73  \\
          & Best  & 77.14  & 74.35  & 72.22  & 68.28  & 59.65  & 63.45  & 53.45  & 49.49  & 45.53  & 40.76  & 60.43  \\
    \hline
    \multirow{2}[2]{*}{DivideMix~\cite{li2020dividemix}} & Last  & 88.11  & 86.95  & 66.42  & 73.78  & 73.85  & 63.08  & 61.14  & 24.31  & 24.13  & 20.09  & 58.19  \\
          & Best  & 88.50  & 87.13  & 67.67  & 74.26  & 74.15  & 63.08  & 61.61  & 33.79  & 31.06  & 26.10  & 60.74  \\
    \hline
    \multirow{2}[2]{*}{ELR+~\cite{liu2020early}} & Last  & 85.21  & 85.71  & 84.13  & 81.16  & 78.91  & 64.76  & 61.41  & 56.42  & 51.31  & 44.05  & 69.31  \\
          & Best  & 87.04  & 86.14  & 84.38  & 83.30  & 80.23  & 65.73  & 62.12  & 56.42  & 52.26  & 45.73  & 70.34  \\
    \hline
    \multirow{2}[1]{*}{MOIT+~\cite{ortego2021multi}} & Last  & 86.67  & 84.60  & 84.85  & 81.34  & 79.02  & 66.21  & 61.63  & 60.01  & 37.71  & 39.01  & 68.11  \\
          & Best  & 87.00  & 84.84  & 84.85  & 81.95  & 79.52  & 66.92  & 62.16  & 60.49  & 37.85  & 39.19  & 68.48  \\
    \hline
    \multirow{2}[2]{*}{Balanced-Softmax~\cite{ren2020balanced}} & Last  & 87.52  & 84.00  & 79.49  & 73.53  & 65.27  & 76.14  & 69.37  & 62.03  & 52.79  & 46.44  & 69.66  \\
          & Best  & 87.62  & 84.50  & 82.84  & 79.09  & 75.90  & 78.63  & 73.29  & 70.89  & 67.57  & 63.40  & 76.37  \\
    \hline
    \multirow{1}[1]{*}{DivideMix} & Last  & 88.23  & 86.96  & 78.02  & 79.17  & 76.89  & 76.63  & 75.72  & 46.82  & 53.25  & 54.33  & 71.60  \\
        +Balanced-Softmax  & Best  & \textbf{88.93}  & 86.06  & 79.19  & 79.81  & 78.20  & 79.36  & 76.55  & 48.60  & 54.99  & 55.84  & 72.75  \\
    \hline
    \multirow{2}[2]{*}{FaMUS~\cite{xu2021faster}} & Last  & 77.31  & 78.68  & 79.15  & 81.17  & 77.57  & 55.46  & 51.67  & 49.40  & 38.21  & 26.69  & 61.53  \\
          & Best  & 83.12  & 84.34  & 84.85  & 85.44  & \textbf{84.12 } & 57.50  & 54.61  & 54.30  & 45.35  & 32.93  & 66.66  \\
    \hline
    \multirow{2}[2]{*}{CurveNet~\cite{jiang2021delving}} & Last  & 84.10  & 81.70  & 78.47  & 78.73  & 75.65  & 65.77  & 66.21  & 62.37  & 48.71  & 51.85  & 69.36  \\
          & Best  & 84.87  & 84.62  & 79.98  & 81.33  & 78.37  & 67.55  & 68.72  & 63.71  & 51.63  & 52.84  & 71.36  \\
    \hline
    \multirow{2}[2]{*}{HAR~\cite{cao2020heteroskedastic}} & Last  & 86.46  & 84.27  & 81.78  & 79.55  & 78.07  & \textbf{78.60}  & {75.05}  & 72.08  & 65.48  & 63.90  & 76.52  \\
          & Best  & 87.03  & 84.47  & 81.94  & 79.87  & 78.25  & \bd{79.02}  & 76.14  & 72.74  & 67.22  & 65.00  & 77.17  \\
    \midrule
    \multirow{2}[2]{*}{\textbf{Dynamic Loss (Ours)} } & Last  & \textbf{89.23 } & \textbf{88.39 } & \textbf{86.58 } & \textbf{84.43 } & \textbf{83.34 } & 77.80  & \textbf{76.31 } & \textbf{74.10 } & \textbf{69.64 } & \textbf{67.45 } & \textbf{79.73 } \\
          & Best   & \textbf{89.44}  & \textbf{88.46 } & \textbf{86.72 } & \textbf{84.73 } & \textbf{83.71}  & 78.96  & \textbf{76.64 } & \textbf{76.17 } & \textbf{70.37 } & \textbf{70.26 } & \textbf{80.55 } \\
    \bottomrule
    \end{tabular}%
    }
\end{table*}%

\begin{table*}[h]
  \centering
  \setlength{\tabcolsep}{6.0pt}
    \renewcommand\arraystretch{1.1}
    \caption{{Accuracy (\%) on CIFAR-100-N-LT with varying imbalance ratio and noise rate. NC: not converging, NA: not available.}}
  \label{tab:sota-cifar100}%
  \resizebox{\textwidth}{!}{
      \begin{tabular}{lc|ccccc|ccccc|c}
    \toprule
    \multicolumn{2}{c|}{\textbf{Imbalance Ratio}} & \multicolumn{5}{c|}{\textbf{10}}               & \multicolumn{5}{c|}{\textbf{100}}              & \multirow{2}[4]{*}{\textbf{Average}} \\
\cmidrule{1-12}    \multicolumn{2}{c|}{\textbf{Noise Rate}}   & \textbf{0.1}   & \textbf{0.2}   & \textbf{0.3}   & \textbf{0.4}   & \textbf{0.5}   & \textbf{0.1}   & \textbf{0.2}   & \textbf{0.3}   & \textbf{0.4}   & \textbf{0.5}   &  \\
    \midrule
    \multirow{2}[2]{*}{Cross Entopy} & Last  & 43.48  & 37.25  & 31.34  & 25.53  & 19.45  & 29.92  & 21.85  & 19.32  & 13.71  & 12.21  & 25.41  \\
          & Best  & 43.95  & 37.64  & 32.18  & 29.44  & 23.87  & 30.52  & 22.13  & 19.58  & 14.59  & 12.81  & 26.67  \\
    \hline
    \multirow{2}[2]{*}{DivideMix~\cite{li2020dividemix}} & Last  & 54.17  & 51.92  & 50.44  & 45.02  & 43.43  & 36.31  & 35.68  & 34.10  & 33.19  & 27.22  & 41.15  \\
          & Best  & 54.94  & 53.35  & 50.93  & 45.36  & 43.44  & 36.99  & 36.24  & 34.87  & 33.64  & 27.74  & 41.75  \\
    \hline
    \multirow{2}[2]{*}{ELR+~\cite{liu2020early}} & Last  & 52.48  & 51.30  & 46.24  & 39.98  & 34.91  & 33.01  & 28.10  & 24.92  & 22.11  & 16.54  & 34.96  \\
          & Best  & 53.91  & 51.90  & 47.88  & 42.61  & 37.35  & 33.81  & 28.94  & 26.10  & 22.11  & 17.39  & 36.20  \\
    \hline
    \multirow{2}[0]{*}{MOIT+~\cite{ortego2021multi}} & Last  & 44.66  & 40.12  & NC    & NC    & NC    & NC    & NC    & NC    & NC    & NC    & NA \\
          & Best  & 47.70  & 44.94  & 42.10  & 39.12  & 35.50  & 32.66  & 30.35  & 28.99  & 25.99  & 22.82  & 35.02  \\
    \hline
    \multirow{2}[2]{*}{Balanced-Softmax~\cite{ren2020balanced}} & Last  & 58.38  & 54.59  & 50.49  & 44.83  & 40.45  & 43.17  & 38.67  & 33.27  & 27.08  & 22.10  & 41.30  \\
          & Best  & 58.62  & 54.73  & 50.66  & 45.63  & 40.56  & 43.50  & 38.67  & 33.62  & 28.05  & 24.19  & 41.82  \\
    \hline
    \multirow{1}[1]{*}{DivideMix} & Last  & 56.37  & 54.80  & 54.83  & 51.29  & 48.89  & 43.26  & 42.42  & 40.46  & 37.83  & 30.95  & 46.11  \\
        +Balanced-Softmax   & Best  & 56.85  & 56.06  & 55.64  & 52.30  & 50.01  & 43.67  & 42.79  & 40.99  & 38.50  & 32.38  & 46.92  \\
    \hline
    \multirow{2}[2]{*}{FaMUS~\cite{xu2021faster}} & Last  & 46.07  & 51.59  & 46.07  & 46.93  & 43.83  & 29.33  & 30.22  & 28.53  & 27.83  & 24.57  & 30.72  \\
          & Best  & 47.03  & 52.05  & 46.41  & 47.88  & 44.30  & 29.66  & 30.31  & 28.50  & 27.24  & 24.85  & 30.81  \\
    \hline
    \multirow{2}[2]{*}{CurveNet~\cite{jiang2021delving}} & Last  & 50.41  & 47.14  & 43.18  & 41.23  & 34.85  & 22.10  & 20.44  & 17.80  & 11.87  & 9.24  & 29.83  \\
          & Best  & 52.73  & 51.93  & 47.56  & 44.08  & 39.74  & 25.26  & 21.35  & 18.72  & 13.60  & 12.20  & 32.72  \\
    \hline
    \multirow{2}[2]{*}{HAR~\cite{cao2020heteroskedastic}} & Last  & 58.88  & 55.43  & 52.57  & 46.01  & 43.96  & 42.67  & 39.39  & 34.43  & 29.43  & 24.94  & 42.77  \\
          & Best  & 59.32  & 55.80  & 53.44  & 46.75  & 44.61  & 44.45  & 40.98  & 36.09  & 31.17  & 27.15  & 43.98  \\
    \midrule
    \multirow{2}[2]{*}{\textbf{Dynamic Loss (Ours)}} & Last  & \textbf{59.24 } & \textbf{57.57 } & \textbf{56.85 } & \textbf{52.07 } & \textbf{50.74 } & \textbf{47.23 } & \textbf{45.74 } & \textbf{42.72 } & \textbf{39.58 } & \textbf{33.87 } & \textbf{48.56 } \\
          & Best  & \textbf{59.52 } & \textbf{57.85 } & \textbf{57.32 } & \textbf{52.66 } & \textbf{51.26 } & \textbf{47.55 } & \textbf{45.82 } & \textbf{43.54 } & \textbf{39.96 } & \textbf{34.30 } & \textbf{48.98 } \\
    \bottomrule
    \end{tabular}%
    }
\end{table*}%

\section{Experiments on Long-Tailed Noisy Data}

\subsection{Experiments on CIFAR-N-LT}
\noindent\textbf{Dataset and Implement Details.}
{We assess the performance of our method on CIFAR-N-LT dataset, which comprises CIFAR-10 and CIFAR-100~\cite{krizhevsky2009learning}. These datasets contain 60,000 RGB images, out of which 50,000 images are used for training and 10,000 images for testing. The images are evenly distributed across 10 and 100 categories, respectively. Additionally, the datasets are subjected to simulated label noise and class imbalance.}

{To simulate a long-tailed dataset, we follow the exponential profile proposed in~\cite{cao2019learning}, where the imbalance ratio $\rho$ leads to an exponential decay in the sample number across different classes. We then inject label noise into the long-tailed dataset to create the training set, with each sample's label independently changed to class $j$ with probability $\frac{N_j}{N}\lambda$, where $N$ is the total number of training samples, $N_j$ is the frequency of class $j$, and $\lambda$ represents the noise rate.}

{Following ROLT~\cite{wei2021robust}, we consider imbalance ratios of $\rho \in \{10,100\}$ and noise rates of $\lambda \in \{0.1,0.2,0.3,0.4,0.5\}$. The ResNet-32~\cite{he2016deep} is adopted as the classifier, and we train all classifiers using balanced-softmax for 300 epochs with a batch size of 512. The learning rate is initialized as $0.1$ and controlled by a cosine annealing learning scheduler~\cite{loshchilov2016sgdr}. We train all classifiers using the same SGD optimizer with a momentum of 0.9 and a weight decay of 5e-4. Further training details can be found in Table~\ref{tab:training_details}.}

\noindent\textbf{Main Results.}
{Table~\ref{tab:sota-cifar10} and ~\ref{tab:sota-cifar100} present the average accuracy on CIFAR-10-N-LT and CIFAR-100-N-LT with varying imbalance ratios and noise rates. Our method exhibits consistently high accuracy across a wide range of biases, whereas previous methods suffer rapid degradation.
Specifically, our dynamic loss improves the average last accuracy by $3.21\%$ and $5.79\%$ compared to HAR on CIFAR-10-N-LT and CIFAR-100-N-LT, respectively. Additionally, our method significantly outperforms the baseline model that simply combines strategies from DivideMix and Balanced-Softmax.}

{Furthermore, the performance of our last model is generally very close to that of our best model despite varying bias settings. In contrast, the last models of DivideMix and Balanced-Softmax degrade significantly compared to their corresponding best models, especially on CIFAR-10-N-LT with severe imbalance and noise (e.g., $\rho=100$ and $\lambda=0.5$). These results suggest that our method is much more resistant to overfitting on biased data than the aforementioned priors.}

{It is worth noting that previous methods require carefully tuned hyperparameters based on unobservable noise rate~\cite{li2020dividemix} or perform two-stage training to obtain prior information on class distribution~\cite{cao2020heteroskedastic}. In comparison, our dynamic loss employs a fixed set of hyperparameters and requires only one-round end-to-end training without manual interventions on the same dataset.}

\begin{table}[t]
  \centering
  \setlength{\tabcolsep}{6.0pt}
      \renewcommand\arraystretch{1.2}
     \caption{{Accuracy (\%) on WebVision and ImageNet validation sets. $^*$ denotes the use of model cotraining or ensembling. $^\dag$ indicates the backbone is pretrained using self-supervised techniques (CLIP). IRV2: Inception-ResNet V2.}}
    \begin{tabular}{l|l|cc|cc}
    \toprule
    \multirow{2}[4]{*}{\textbf{Methods}} & \multirow{2}[4]{*}{\textbf{Backbone}} & \multicolumn{2}{c|}{\textbf{Webvision}} & \multicolumn{2}{c}{\textbf{ImageNet}} \\
\cmidrule{3-6}          &       & \textbf{top 1} & \textbf{top 5} & \textbf{top 1} & \textbf{top 5} \\
    \midrule
    Cross Entropy  & IRV2 &  72.48  & 88.48  & 65.08  & 87.88  \\
    Cross Entropy  & ResNet-50 &  71.72 & 87.84 & 65.96  & 86.88  \\
     Cross Entropy  & ResNet-50$^\dag$ &  78.04 & 93.04 & 72.84 & 91.68  \\
    HAR~\cite{cao2020heteroskedastic}   & IRV2 & 75.50  & 90.70  & 70.30  & 90.00  \\
    DivideMix$^*$~\cite{li2020dividemix} & IRV2 & 77.32  & 91.64  & 75.20  & 90.84  \\
    ROLT+$^*$~\cite{wei2021robust} & IRV2 & 77.64  & 92.44  & 74.64  & 92.48  \\
    ELR+$^*$~\cite{liu2020early}  & IRV2 & 77.78  & 91.68  & 70.29  & 89.76  \\
    CMW-Net-SL$^*$~\cite{shu2022cmw} & IRV2 & 78.08  & 92.96  & \textbf{75.72 } & 92.52  \\
    FaMUS$^*$~\cite{xu2021faster} & IRV2 & 79.40  & 92.80  & 77.00  & 92.76  \\
    MOIT+~\cite{ortego2021multi} & IRV2 & 78.76  & \textbf{-} & -     & \textbf{-} \\
    NCR~\cite{iscen2022learning}   & ResNet-50 & 80.50  & \textbf{-} & -     & \textbf{-} \\
    INOLML$^*$~\cite{hoang2022maximising} & ResNet-50 & 81.70  & 93.80  & \textbf{78.10}  & 92.90  \\
    \midrule
    \textbf{Dynamic Loss}  & IRV2 & 80.12 & 93.64 & 74.76 & 93.08 \\
    \textbf{Dynamic Loss}  & ResNet-50 & 78.56 & 92.52 & 72.08 & 91.36 \\  
    \textbf{Dynamic Loss}  & ResNet-50$^\dag$ &  \textbf{81.96} & \textbf{94.48}    & 77.43 & \textbf{93.40}  \\
    \bottomrule
    \end{tabular}%
  \label{tab:webvision}%
\end{table}%

\subsection{Experiments on Webvision.}

\noindent\textbf{\R{Dataset and Implement Details.}} 
{We also evaluate our dynamic loss on the WebVision dataset~\cite{li2017webvision}, which is a large-scale real-world dataset that suffers from label noise and class imbalance. It comprises 2.4 million images, of which approximately $20\%$ are mislabeled~\cite{chen2021two}. To construct the miniWebVision dataset, we follow the methodology proposed in MentorNet~\cite{jiang2018mentornet} by selecting the top $50$ classes, resulting in an observed imbalance ratio of around $6.78$. 
Following priors~\cite{li2020dividemix}, we train the Inception-ResNet V2~\cite{szegedy2017inception} for $150$ epochs using the SGD optimizer with momentum $0.9$ and weight decay of 1e-4.
For ResNet-50, the training details is following INOLML~\cite{hoang2022maximising}.
During the warm-up stage, which lasts for one epoch, we use a batch size of 64 and an initial learning rate of $0.02$. The learning rate is then adjusted using a cosine annealing learning scheduler. We evaluate the model's performance on the validation sets of both WebVision and ImageNet~\cite{deng2009imagenet}. Additional training details are available in Table~\ref{tab:training_details}.}

\noindent\textbf{Main Results.}  
{Table~\ref{tab:webvision} displays the performance of our method on the WebVision and ImageNet validation sets. Our approach outperforms other state-of-the-art methods even though most of them utilize additional model cotraining and ensembling techniques. Notably, our method surpasses HAR and ROLT+, which are specifically designed to handle long-tailed noisy data, by at least $2.48\%$ in terms of accuracy on WebVision, demonstrating its superiority.}

{The rapid advancements in the field of self-supervised learning have highlighted the increasing importance of adaptive methods for learning from biased data. Particularly, methods that can effectively adapt to models with varying initial parameters are of great significance. In this study, we performed experiments on ResNet-50 that was pre-trained using self-supervised techniques, such as CLIP~\cite{DBLP:conf/icml/RadfordKHRGASAM21}. The results, presented in the last row of Table~\ref{tab:webvision}, demonstrate a significant improvement of 3.92\% in accuracy, achieved through our proposed approach, compared to the model trained using cross entropy with the same initial parameters pretrained by CLIP. These findings provide evidence of the dynamic adaptability of our method to classifiers with different initial parameters.}

\begin{table*}[h]
  \centering
  \setlength{\tabcolsep}{8.8pt}
  \renewcommand\arraystretch{1.2}
  \caption{{Accuracy (\%) on CIFAR-N with varying noise rate. PARes18 used as the classifier.}}
  \label{tab:noise_only_cifar}%
    \begin{tabular}{l|ccccc|c|ccc|c}
    \toprule
    \multicolumn{1}{c|}{\textbf{Datasets}} & \multicolumn{6}{c|}{\textbf{CIFAR-10-N} }                & \multicolumn{4}{c}{\textbf{CIFAR-100-N}} \\
    \midrule
    \multicolumn{1}{c|}{\textbf{Noise Rate}} & \textbf{20}    & \textbf{40}    & \textbf{60}    & \textbf{20 (Asym.)} & \textbf{40 (Asym.)} & \textbf{Avg.} & \textbf{20}    & \textbf{40}    & \textbf{60}    & \textbf{Avg.} \\
    \midrule
    Cross Entopy    & 86.98  & 77.52  & 73.63  & 83.60  & 77.85  & 79.92  & 60.38  & 46.92  & 31.82  & 46.37  \\
    SELFIE~\cite{song2019selfie} & 86.39  & 82.23  & 74.81  & -     & -     & -     & 55.71  & 51.14  & 43.85  & 50.23  \\
    PLC~\cite{prog_noise_iclr2021}   & 86.40  & 71.72  & 65.22  & 90.23  & 85.40  & 79.79  & 59.66  & 49.24  & 33.18  & 47.36  \\
    NCT~\cite{chen2022compressing}    & 95.00  & 87.00  & 73.22  & 91.51  & 93.00  & 87.95  & 67.65  & 57.97  & 45.01  & 56.88  \\
    Coteaching~\cite{han2018co} & 93.83  & 91.74  & 57.65  & 93.23  & 90.78  & 85.45  & 70.81  & 62.65  & 41.55  & 58.34  \\
    CMW-Net~\cite{shu2022cmw} & 91.09  & 86.91  & 83.33  & 93.02  & 92.70  & 89.41  & 70.11  & 65.84  & 56.93  & 64.29  \\
    DivdeMix~\cite{li2020dividemix} & 95.63  & 93.78  & \textbf{94.23 } & 94.18  & 92.73  & 94.11  & 77.20  & 73.37  & \textbf{70.75 } & 73.77  \\
    GJS~\cite{englesson2021generalized}   & 94.20  & 92.80  & 89.72  & 91.92  & 86.07  & 90.94  & 73.31  & 71.33  & 66.92  & 70.52  \\
    NCR~\cite{iscen2022learning}   & 95.20  & 94.50  & 78.45  & -     & 90.70  & -     & 76.60  & 74.20  & 38.25  & 63.02    \\
    MOIT+~\cite{ortego2021multi}   & 94.08  & 91.95  & 89.38  & 94.50  & 93.27  & 92.64 & 75.89  & 70.88  & 65.30  & 70.69  \\
    \midrule
    \textbf{Dynamic Loss (Ours)}  & \textbf{95.90 } & \textbf{94.69 } & 92.28  & \textbf{95.74 } & \textbf{94.51 } & \textbf{94.62 } & \textbf{78.26 } & \textbf{75.28 } & 69.18  & \textbf{74.24 } \\
    \bottomrule
    \end{tabular}%
\end{table*}%

\begin{table*}[t]
  \begin{center}  
  \setlength{\tabcolsep}{4.0pt}
  \renewcommand\arraystretch{1.2}
  \caption{{Accuracy (\%) on Animal-10N.}}
  \label{tab:animal10n}%
    \begin{tabular}{c|cccccccc|c}
    \toprule
    \textbf{Methods} & Cross Entropy & SELFIE~\cite{song2019selfie} & PLC~\cite{prog_noise_iclr2021} & NCT~\cite{chen2022compressing} & Co-teaching~\cite{han2018co} & CMW-Net~\cite{shu2022cmw} & DivdeMix~\cite{li2020dividemix} & GJS~\cite{englesson2021generalized} & {\textbf{Dynamic Loss}} \\    
    \midrule
    \textbf{Accuracy} & 79.40 & 81.80 & 83.40 & 84.10  & 80.20 & 84.70 & 84.00  & 84.17 & \bd{86.54} \\
    \bottomrule
    \end{tabular}%
  \end{center}
\end{table*}%

\section{Experiments on Noisy Data}
\subsection{Experiments on CIFAR-N}
\noindent\textbf{Dataset and Implement Details.}
{CIFAR-N is a synthetic noisy dataset derived from CIFAR. It includes two common types of simulated label noise: symmetric and asymmetric. Symmetric noise is introduced by randomly changing the labels with all possible labels based on a fixed probability of $\lambda$ (also known as noise rate). Asymmetric noise, on the other hand, is designed manually to mimic real-world label noise, where labels are only altered with those in similar classes, such as deer $\rightarrow$ horse and dog $\leftrightarrow$ cat. We evaluate the performance of our dynamic loss in handling noisy data using CIFAR-10-N and CIFAR-100-N with symmetric noise rates of ${0.2, 0.4, 0.6}$ and asymmetric noise rates of ${0.2, 0.4}$. We utilize PreAct ResNet (PARes18)\cite{he2016identity} as the classifier, following the approach proposed in DivideMix\cite{li2020dividemix}. Additional training details can be found in Table~\ref{tab:training_details}.}

\noindent\textbf{Results on CIFAR-N.} 
{Table~\ref{tab:noise_only_cifar} illustrates that our proposed method outperforms previous methods specifically designed for learning on noisy data, achieving the highest average accuracy. Unlike DivideMix, which requires manual tuning of hyperparameters under different noise types and rates, our dynamic loss adapts well to various noisy scenarios in a fully self-adaptive manner without any manual intervention.}

\subsection{Experiments on Animal-10N}
\noindent\textbf{Dataset and Implement Details.}
{We also evaluate our method on the real-world noisy dataset ANIMAL-10N~\cite{song2019selfie}, which contains a total of 55,000 images (50,000 for training and 5,000 for testing) of 5 pairs of confusing animals. All image categories are equally represented in the dataset. The images are collected from websites using the search keywords as labels, resulting in significant label noise with an estimated rate of 8\%. To ensure a fair comparison with prior work~\cite{song2019selfie}, we train the VGG10-BN~\cite{simonyan2014very} for $100$ epochs using the SGD optimizer with a momentum of $0.9$ and weight decay of 5e-4. The warm-up stage lasts for 5 epochs, and the batch size is set to 128. We initialize the learning rate to 0.02 and control it with a cosine annealing learning scheduler. Further training details are presented in Table~\ref{tab:training_details}.}

\noindent\textbf{Main Results.} 
{Table~\ref{tab:animal10n} illustrates that our dynamic loss achieves state-of-the-art performance compared to all previous methods on the ANIMAL-10N dataset. Specifically, our method outperforms the classifier trained with cross entropy by a large margin of 7.14\%, demonstrating its superior ability to handle real-world noisy data.}

\section{Experiments on Long-Tailed Data}
\subsection{Experiments on CIFAR-LT}
\noindent\textbf{Dataset and Implement Details.}
{The CIFAR-LT is a simulated long-tailed dataset that is derived from CIFAR by reducing the number of training samples per class according to an exponential function $n_i = n \mu^i$, where $n_i$, $i$, and $n$ denote the number of samples in the $i$-th class, the class index, and the maximum number of samples across all classes, respectively. We evaluate the efficacy of our dynamic loss approach in dealing with long-tailed data on clean CIFAR datasets that have varying imbalance ratios ($\rho \in \{10,20,50,100\}$). For more detailed information on the training process, please refer to Table~\ref{tab:training_details}.}

\begin{table*}[h]
  \centering
  \setlength{\tabcolsep}{9.0pt}
  \renewcommand\arraystretch{1.2}
  \caption{{Accuracy (\%) on CIFAR-LT with varying imbalance ratio. ResNet32 used as the classifier.}}
  \label{tab:imb_only_ciafr}%
    \begin{tabular}{l|cccc|c|cccc|c}
    \toprule
    \multicolumn{1}{c|}{\textbf{Datasets}} & \multicolumn{5}{c|}{\textbf{CIFAR-10-LT}}      & \multicolumn{5}{c}{\textbf{CIFAR-100-LT}} \\
    \midrule
    \multicolumn{1}{c|}{\textbf{Imbalance Ratio}} & \textbf{10}    & \textbf{20}    & \textbf{50}    & \textbf{100}   & \textbf{Avg.} & \textbf{10}    & \textbf{20}    & \textbf{50}    & \textbf{100}   & \textbf{Avg.} \\
    \midrule
    Cross entropy & 86.39 & 82.23 & 74.81 & 70.36 & 78.45  & 55.71 & 51.14 & 43.85 & 38.32 & 47.26  \\
    Focal Loss~\cite{lin2017focal} & 86.66 & 82.76 & 76.71 & 70.38 & 79.13  & 55.78 & 51.95 & 44.32 & 38.41 & 47.62  \\
    CB Focal~\cite{cui2019class} & 87.49 & 84.36 & 79.27 & 74.57 & 81.42  & 57.99 & 52.59 & 45.32 & 39.6  & 48.88  \\
    LDAM-DRW~\cite{cao2019learning} & 87.68 & 85.51 & 81.64 & 78.02 & 83.21  & 44.70  & 52.93 & 48.22 & 59.59 & 51.36  \\
    FaMUS~\cite{xu2021faster} & 87.9  & 86.24 & 83.32 & 80.96 & 84.61  & 59.00    & 55.95 & 49.93 & 46.03 & 52.73  \\
    Balanced-Softmax~\cite{ren2020balanced} & 91.01  & 88.85  & 86.44  & 82.31  & 87.15  & 64.00  & 59.48  & 54.36  & 50.47  & 57.08  \\
    WD~\cite{alshammari2022long} &89.80  & 84.81 & 79.66  & 74.84 & 82.28  & 61.60  & 52.75 & 45.89 & 40.79 & 50.26   \\
    MiSLAS~\cite{zhong2021improving} & 90.00  & 88.52  & 85.70  & 82.10  & 86.58  & 63.20  & 59.25 & 52.30  & 47.00  & 55.44  \\
    Logit Adjustment~\cite{menonlong} & 89.64 & 86.77 & 82.61 & 78.38 & 84.35  & 62.83 & 58.81  & 52.15 & 48.36 & 55.54  \\
    CMO~\cite{park2022majority}   & 83.26  & 89.27  & \textbf{87.19}  & \textbf{85.35}  & 86.27  & 62.30  & 60.12  & 51.40  & 46.60  & 55.11  \\
    \midrule
    \textbf{Dynamic Loss}  & \textbf{91.24 } & 88.30  & 86.46 & 82.95 & 87.24 & 63.99  & 59.79  & 54.51  & 50.14  & 57.11 \\
    \textbf{Dynamic Loss + Balanced-Softmax} & 90.99 & \textbf{89.66} & 85.49 & 83.21 & 87.34  & \textbf{64.18}  & \textbf{60.13}  & \textbf{54.60}  & \textbf{50.54} & \textbf{57.36}  \\
    \textbf{Dynamic Loss + Logit Adjustment} & 91.10  & 89.54 & 85.52 & 83.43 & \textbf{87.40  } & 64.15 &  60.08 & 54.14 & 49.65 & 57.01  \\
    \bottomrule
    \end{tabular}%
\end{table*}%

\noindent\textbf{Results} 
{Table~\ref{tab:imb_only_ciafr} presents the superior performance of our proposed method compared to previous approaches specifically designed for learning on long-tailed data. Notably, our method surpasses LDAM, which adjusts classification margins based solely on the sample number, by a significant margin of $4.03\%$ and $5.75\%$ on CIFAR-10-LT and CIFAR-100-LT, respectively. This result demonstrates the effectiveness of our dynamic loss in accurately perceiving the classification difficulty of different classes and adaptively adjusting their margins accordingly.}

{Furthermore, our approach seamlessly integrates with Logit Adjustment and Balanced Softmax techniques by utilizing them as base margins while learning the residual margin through our margin generator. As shown in Table~\ref{tab:imb_only_ciafr}, our proposed approach with Balanced-softmax and logit adjustment yields a notable improvement in final accuracy, namely 0.19\% and 3.05\% on CIFAR-10-LT, compared to the initial Balanced-softmax and logit adjustment techniques. However, the performance level remains comparable to that of our original dynamic loss. These findings indicate that our original dynamic loss is sufficient for accurately identifying the classification complexity of diverse categories and accounting for the number of samples per class.}

\subsection{Experiments on ImageNet-LT}
\noindent\textbf{Dataset and Implement Details.}
{The ImageNet-LT dataset consists of 115.8K images, categorized into 1,000 classes based on the Pareto distribution. Consequently, the number of images per class varies between 5 and 1280. Following previous studies~\cite{liu2019large, cui2021parametric}, we employ several architectures for training, including ResNet-10, ResNet-50, ResNet-152, ResNeXt-50~\cite{xie2017aggregated}, and self-supervised ResNet-50. These models are trained for either 90 or 400 epochs, utilizing the SGD optimizer with a momentum of 0.9 and weight decay of 1e-4. No warm-up stage is conducted, and a batch size of 128 is employed. The learning rate is initialized at 0.1 and controlled using a cosine annealing learning scheduler. Further details regarding the training procedure are provided in Table~\ref{tab:training_details}.}

\noindent\textbf{Main Results}
{Table~\ref{tab:imagenet-lt} presents the experimental findings, showcasing the superior accuracy achieved by our proposed method when employed in conjunction with various classifiers, namely ResNet-50, ResNet-152, ResNeXt-50, and self-supervised ResNet-50. Remarkably, even when utilizing the ResNet-10 classifier trained from scratch, our method surpasses the performance of the PaCo framework~\cite{cui2021parametric} equipped with self-supervised techniques by 2.12\%. These results underscore the robustness and efficacy of our proposed method in effectively handling the challenges posed by long-tailed datasets.}

{Additionally, we conducted supplementary experiments by incorporating our method with the pretrained ResNet50 using self-supervised techniques. The results reveal a significant performance improvement of 2.42\% compared to the PaCo framework. This further highlights the compatibility of our proposed method with self-supervised approaches, leading to enhanced performance gains.}

\begin{table}[t]
  \begin{center}  
  \setlength{\tabcolsep}{3.0pt}
  \renewcommand\arraystretch{1.2}
  \caption{{Accuracy (\%) on Imagenet-LT. * indicates the use of self-supervised techniques. RN: ResNet. RNeXt: ResNeXt.}}
  \label{tab:imagenet-lt}%
   \begin{tabular}{l|c|cccc}
    \toprule
    \textbf{Methods} & \textbf{Epoch} & \textbf{RN-10} & \textbf{RN-50} & \textbf{RN-152} & \textbf{RNeXt-50} \\
    \midrule
    Cross Entopy & 90    & 34.01  & 44.60  & 46.20  & 42.78  \\
    Focal Loss~\cite{lin2017focal} & 90    & 32.64  & 41.61  & 44.36  & 41.58  \\
    LDAM-DRW~\cite{cao2019learning} & 90    & 36.03  & 48.80  & 51.83  & 51.43  \\
    CDB-S~\cite{sinha2022class} & 90    & 37.70  & 41.80  & 46.40  & 45.10  \\
    Logit adjustment~\cite{menonlong}  & 90    & 38.43  & 48.89  & 47.86  & 51.85  \\
    Balanced-Softmax~\cite{ren2020balanced}  & 90    & 38.21  & 50.96  & 53.93  & 51.73  \\
    MiSLAS~\cite{zhong2021improving} & 400   & 44.36  & 53.05  & 48.82  & 51.88  \\
    PaCo$^*$~\cite{cui2021parametric}  & 400   & 42.89  & 57.00  &   \textbf{58.55}    & \textbf{58.20}  \\
    \midrule
    \textbf{Dynamic Loss}  & 90    &    38.87   &    51.16   &   54.40    &  52.37 \\
    \textbf{Dynamic Loss}  & 400   &   \textbf{45.01}    &   53.19    &   56.41   & 53.48  \\
    \textbf{Dynamic Loss}$^*$& 400   &   -    &   \textbf{59.42}   &   -   & -  \\
    \bottomrule
    \end{tabular}%
  \end{center}
\end{table}%

\begin{figure*}[t]
 \centering
    \includegraphics[width=\linewidth]{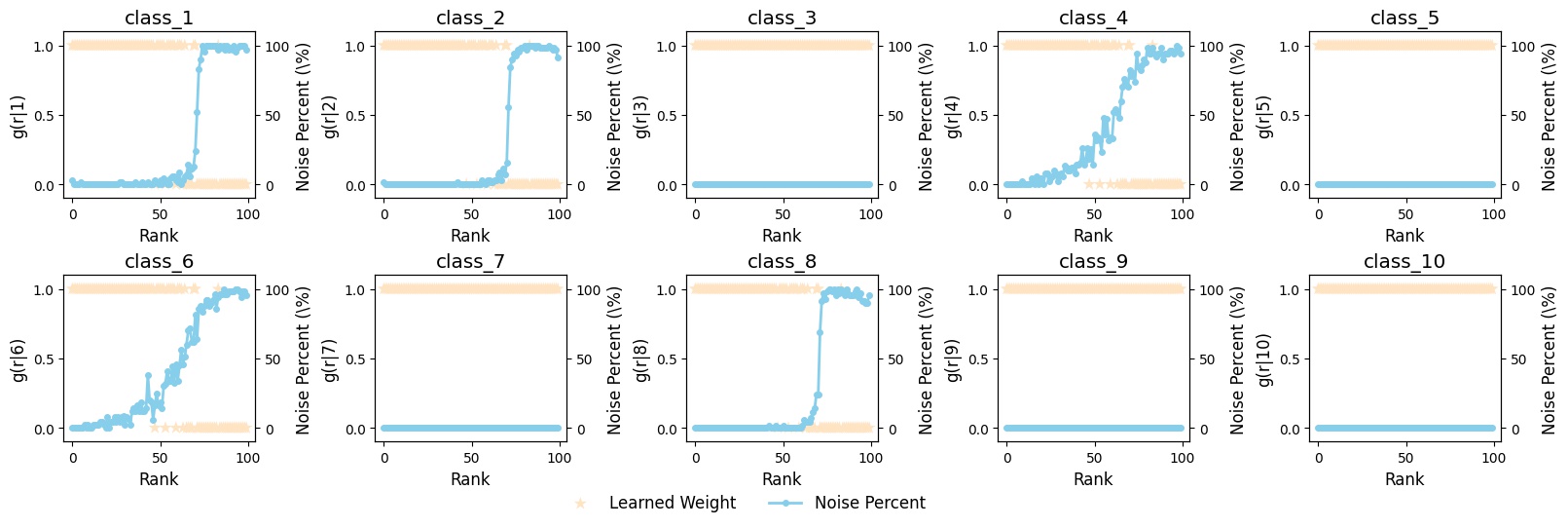}
    \caption{{Visualizing the learned label weights $g(r|y)$ and noise percentage for each class on CIFAR-10-N (Asym. $\lambda=0.4$).}}
    \label{fig:vis:asym_40}
\end{figure*}

\begin{figure}[t]
 \centering
    \includegraphics[width=\linewidth]{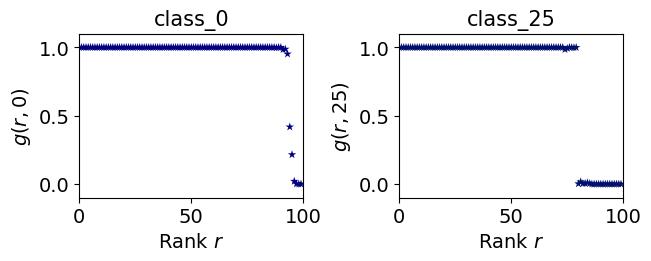}
    \caption{{The visualization of the per-class learned label weights on WebVision.}}
    \label{fig:vis:webvision_rank}
\end{figure}

\begin{figure}[t]
 \centering
    \includegraphics[width=\linewidth]{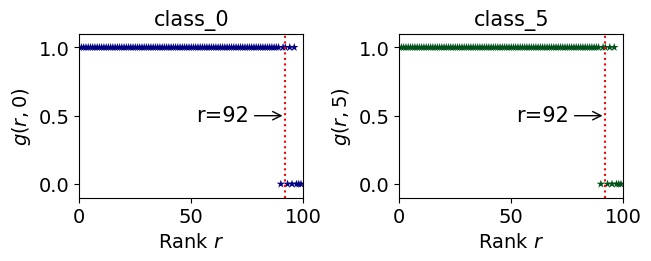}
    \caption{{The visualization of the per-class learned label weights on Animal-10N.}}
    \label{fig:vis:animal_rank_class}
\end{figure}

\section{Qualitative Analysis.}
\subsection{Label corrector}
\noindent\textbf{Behavior of Label Corrector.}
{We examine the behavior of the label corrector on the balanced CIFAR-10-N with asymmetric noise, which is designed to emulate the structure of real-world label noise by assigning distinct noise rates to various classes. Figure~\ref{fig:vis:asym_40} illustrates the learned weight $g(r|y)$ by the label corrector and the percentage of noisy labels corresponding to an increasing loss bin index $r$ for each class. We observe that for the classes that contain noisy labels, clean samples mainly appear in the top-ranked (low-loss) bins while noisy samples occupy the bottom-ranked (high-loss) bins. This finding supports our hypothesis that the loss bin index $r$ can be used as a reliable input indicator for the label corrector to differentiate between noisy and clean samples. Accordingly, the generated weight $g(r|y)$ remains at $1$ and suddenly drops to $0$ at around bin $60$, indicating that the label corrector preserves the assigned ground-truth label for clean samples and resorts to the predicted label that is more likely to be the ground-truth for noisy samples. On the other hand, for the classes without noisy labels, $g(r|y)$ remains at $1$. As a result, our label corrector consistently outputs the correct labels for both noisy and clean samples across different classes.}

{We also visualize the learned label weights of the real-world datasets miniWebvision and Animal-10n in Figures~\ref{fig:vis:webvision_rank} and \ref{fig:vis:animal_rank_class}, respectively. For miniWebvision, we select 2 categories at intervals of 25 to visualize their learned label weights considering the large number of categories. As shown in Figure~\ref{fig:vis:webvision_rank}, the learned label weights vary with different classes, suggesting that the noise rates of different classes are different, which is consistent with real-world datasets. In Animal-10N, two categories were selected at intervals of 5 to visualize the learned label weights. As depicted in Figure~\ref{fig:vis:animal_rank_class}, the learned label weights remain $1$ and drop to $0$ at around bin $92$ (red dotted line), indicating that the noise rate estimated by the label corrector is about $8\%$, which is consistent with the well-recognized estimated noise rate on Animal-10N~\cite{song2019selfie}.}

\begin{figure}[t]
 \centering
 \begin{subfigure}[b]{0.48\linewidth}
    \includegraphics[width=\linewidth]{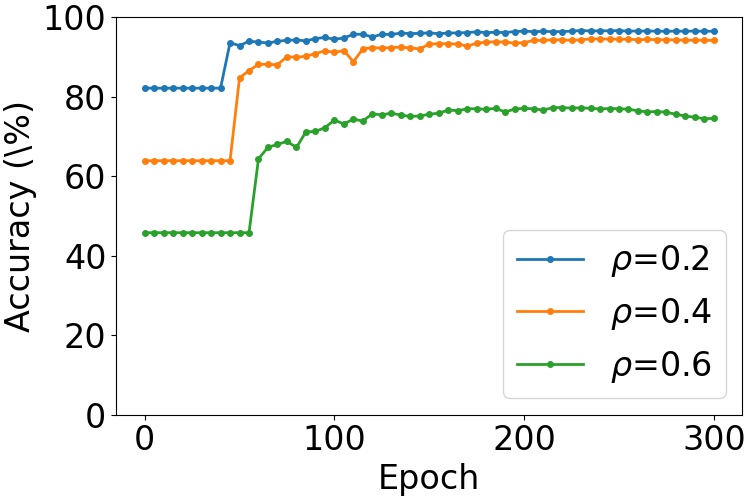}
  \label{fig:vis:cifar10_noise}
  \end{subfigure} \hfill
  \begin{subfigure}[b]{0.48\linewidth}
  \includegraphics[width=\linewidth]{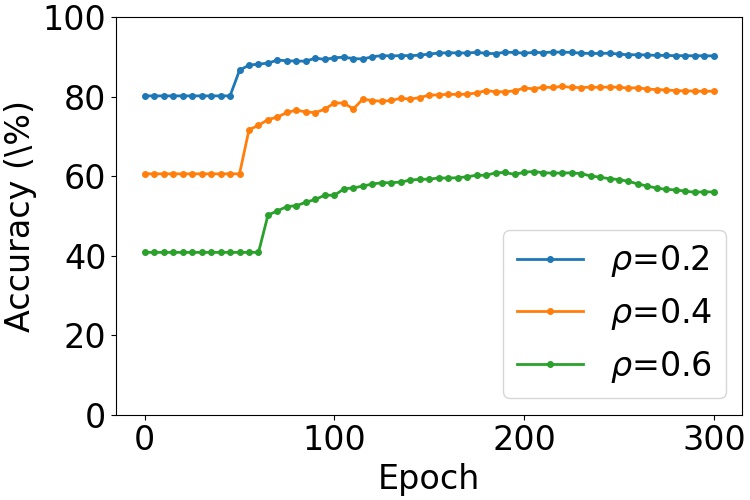}
  \label{fig:vis:cifar100_noise}
  \end{subfigure}
    \caption{{The visualization of the accuracy of generated labels varied with epoch of  $\mathcal{G}_{\bm{\theta}_l}$ on CIFAR-10-N (left) and CIFAR-100-N (right) with noise rates $\lambda$ ranging from 0.2 to 0.6.}}
    \label{fig:vis:noise}
  \end{figure}

\noindent\textbf{Corrected Label Accuracy.}
{Figure~\ref{fig:vis:noise} illustrates the accuracy of the corrected labels on balanced CIFAR-10-N, measured as the proportion of samples with ground-truth labels after the label correction process. The label accuracy gradually increases as the training progresses, and the classifier becomes more reliable. Eventually, the label accuracy reaches over $90\%$ on CIFAR-10-N with noise rates of 0.2 and 0.4. Remarkably, the accuracy has also significantly improved by $35\%$ under a noise rate of $0.6$. The high accuracy of the corrected labels validates our design choices from two perspectives. Firstly, this finding supports our assumption that the classifier primarily focuses on fitting in the dominant clean samples, and can transfer the acquired knowledge to the noisy samples for predicting their ground-truth labels. Secondly, the label corrector can precisely identify the noisy samples and rectify their labels with the predicted correct ones.}

\begin{figure*}[t]
\centering
\begin{subfigure}{.33\textwidth}
  \centering
  \includegraphics[width=\linewidth]{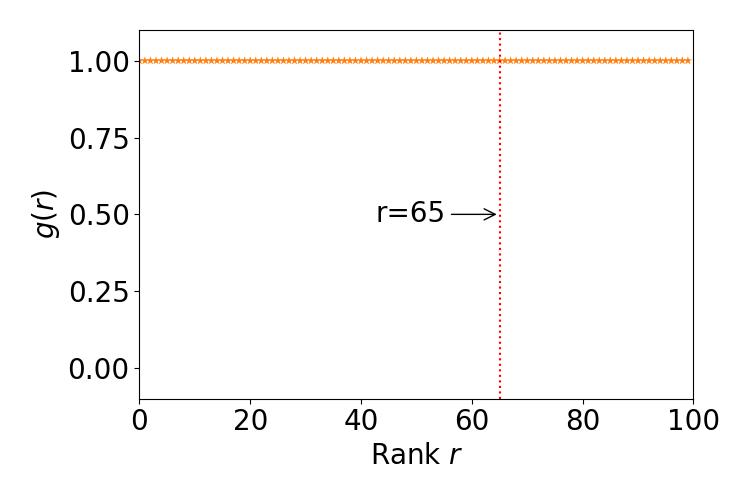}
  \caption{Epoch 10}
  \label{fig:rank_epoch10}
\end{subfigure} \hfill%
\begin{subfigure}{.33\textwidth}
  \centering
  \includegraphics[width=\linewidth]{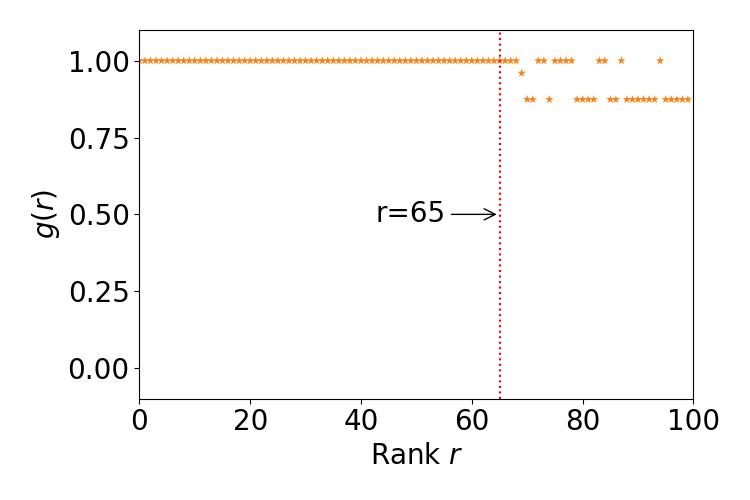}
  \caption{Epoch 45}
  \label{fig:rank_epoch45}
\end{subfigure} \hfill %
\begin{subfigure}{.33\textwidth}
  \centering
  \includegraphics[width=\linewidth]{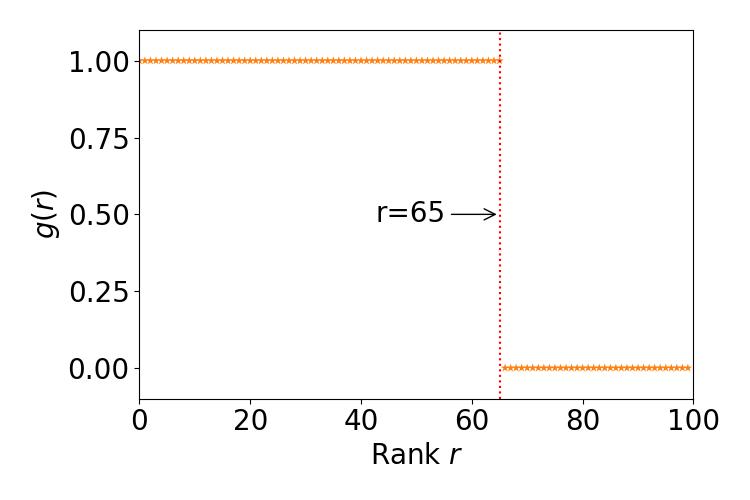}
  \caption{Epoch 50}
  \label{fig:rank_epoch50}
\end{subfigure} 
\caption{{The visualization of the learnable weight $g(r)$ of class 1 varied with training epochs on CIFAR-10-N with noise rate of $\lambda$=0.4.}}
\label{fig:rank_epoch}
\end{figure*}

\noindent\textbf{Learnable Weight Varied Trend in Training.}
{Figure~\ref{fig:rank_epoch} illustrates the evolution of the learnable weight across training epochs.
At the beginning, the label corrector mainly relies on the given labels to train the classifier, while gradually shifting towards trusting the predicted labels for samples with higher rank bins, as indicated by the decreasing values of the learnable weight at epoch 45. Furthermore, the label corrector accurately estimates the noise rate to be approximately 35\% (for a noise rate of 40\%, there are actually 35\% noisy samples). The plot, together with Figure~\ref{fig:vis:noise}, reveals that the epoch at which the label corrector begins to trust the classifier is delayed as the noise rate increases.
This suggests that the label corrector considers the classifier to require more training epochs to produce more reliable predicted labels as the noise rate of the training set increases.
Therefore, our results demonstrate that the label corrector is capable of dynamically adjusting the relabelling of noisy labels based on the status of the classifier and the training set.}

\subsection{Margin generator}
\begin{figure}[t]
\centering
   \includegraphics[width=1.0\linewidth]{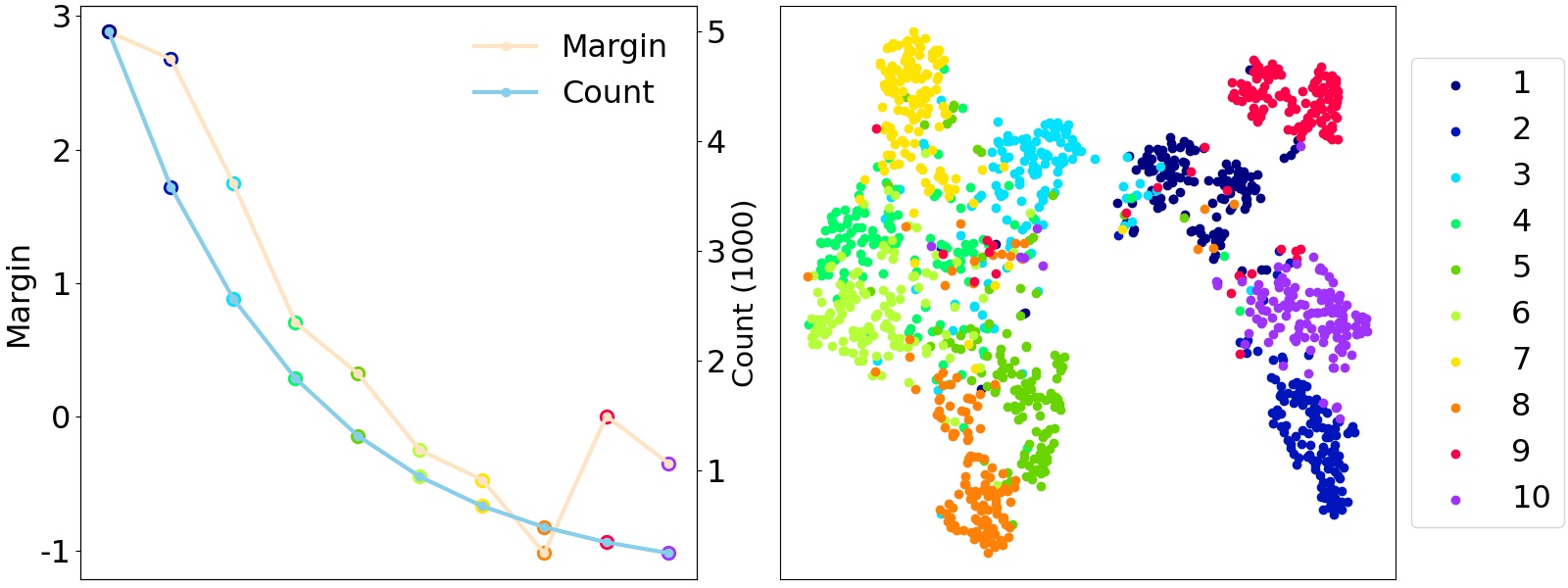}
   \caption{{The learned per-class margins and the number of samples (left), and the feature distribution of samples in the meta set (right) on CIFAR-10-LT with $\rho$=20, are presented.}}
\label{fig:vis:cifar10_margin_tsne}
\end{figure}

\noindent\textbf{Behavior of Margin Generator.}
{We conducted an analysis of the behavior of the margin generator on clean CIFAR-10-LT with imbalance factor $\rho=20$. The left subfigure of Figure~\ref{fig:vis:cifar10_margin_tsne} shows the generated margins for different classes. We observed that as the class index increases and the sample number decreases, the learned margin also decreases as expected. This suggests that the margin generator has the ability to automatically discern the sample numbers of different classes and adaptively adjust the margins for each class accordingly.}

{Interestingly, we also observed some irregularly larger margins on class $9$ and $10$. To explain this observation, we visualized the feature distribution of the meta set using T-SNE~\cite{van2008visualizing} as shown in the right subfigure of Figure~\ref{fig:vis:cifar10_margin_tsne}. We found that the feature distribution of these two classes correspond to the two rightmost clusters, indicating that they are easier to be distinguished from the other classes. This evidence supports the claim that the margin generator takes into account not only the sample number but also the classification difficulty of each class when generating comprehensively adaptive margins during classifier training.}

\begin{figure}[t]
 \centering
   \includegraphics[width=0.85\linewidth]{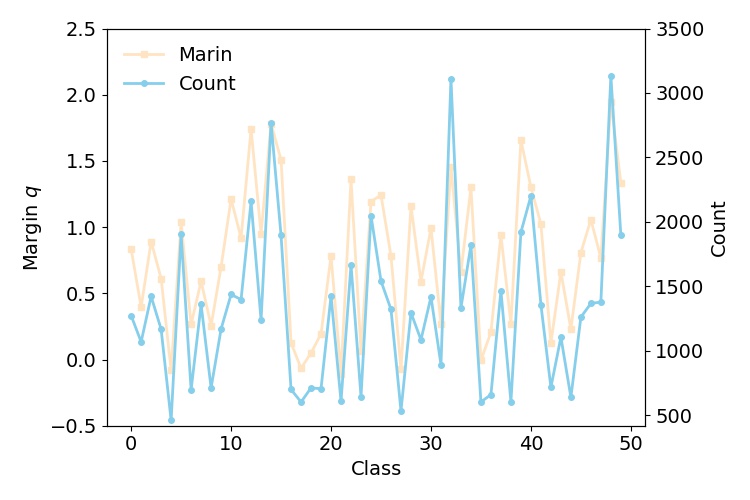}
    \caption{{Visualization of the learned per-class margins on Webvision.}}
    \label{fig:vis:webvision_margin}
   \end{figure}

\begin{figure}[t]
 \centering
    \includegraphics[width=\linewidth]{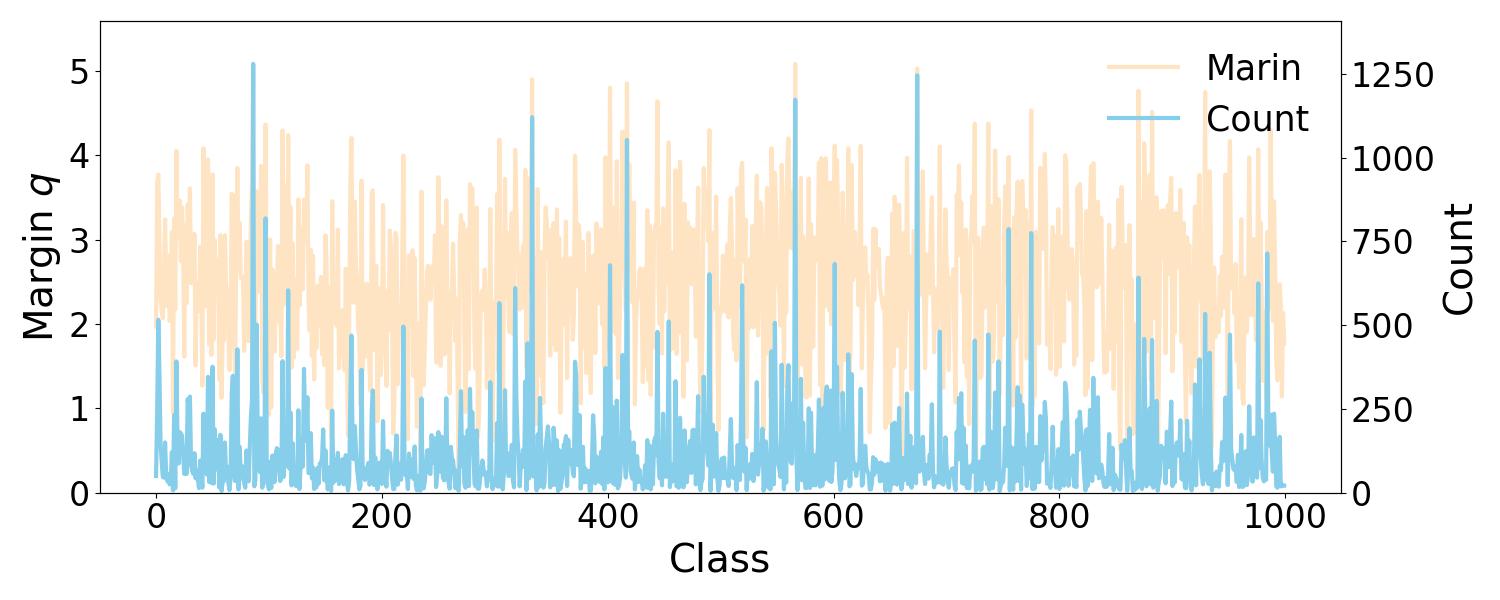}
    \caption{{Visualization of the learned class margins on ImageNet-LT.}}
    \label{fig:vis:im_margin}
\end{figure}

{We present visualizations of the learned label weights for the real-world miniWebvision and simulated ImageNet-LT datasets in Figures~\ref{fig:vis:webvision_margin} and~\ref{fig:vis:im_margin}, respectively.
In Figure~\ref{fig:vis:webvision_margin}, the generated margins over different classes in miniWebvision accord with the complex variation of sample size, which demonstrates the adaptability of our method to handle complex real-world biased data.
For the simulated ImageNet-LT dataset, we visualize the learned margins for 333 categories in intervals of 3 in Figure~\ref{fig:vis:im_margin}.
As shown in the figure, the learned margins consistently vary with the sample number of different classes, suggesting that the margin generator can generate proper margins for different classes.
This finding provides evidence that our method can handle datasets with numerous categories.}

\begin{figure}
 \centering
 \begin{subfigure}[b]{0.48\linewidth}
    \includegraphics[width=\linewidth]{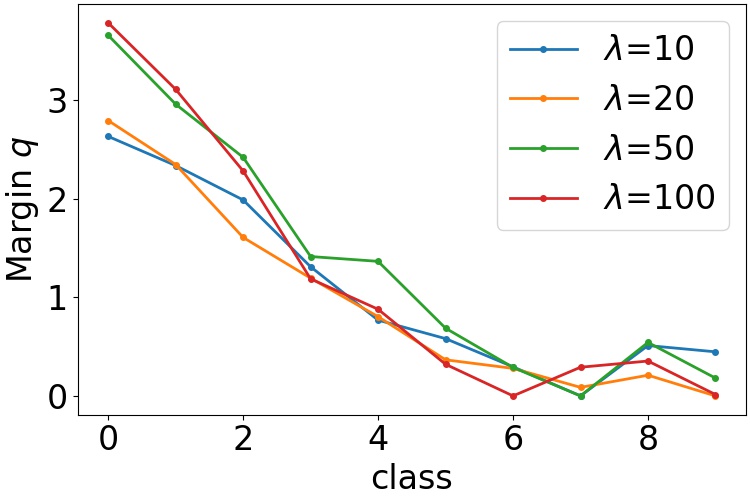}
    \vspace{-0.6cm}
  \label{fig:vis:cifar10_imblance}
  \end{subfigure} \hfill
  \begin{subfigure}[b]{0.48\linewidth}
  \includegraphics[width=\linewidth]{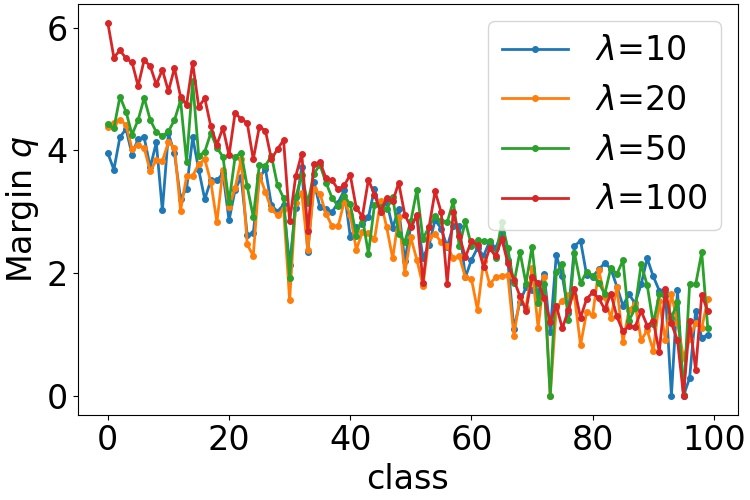}
  \vspace{-0.6cm}
  \label{fig:vis:cifar100_imbalance}
  \end{subfigure}
    \caption{{The visualization of the learned class-aware margin of $\mathcal{G}_l({\bm{\theta}}_l)$ on CIFAR-10-LT (left) and CIFAR-100-LT (right) with imbalance factor $\rho$ ranging from 10 to 100.}}
    \label{fig:vis:imbalance}
  \end{figure}

\begin{figure*}[t]
 \centering
    \includegraphics[width=\linewidth]{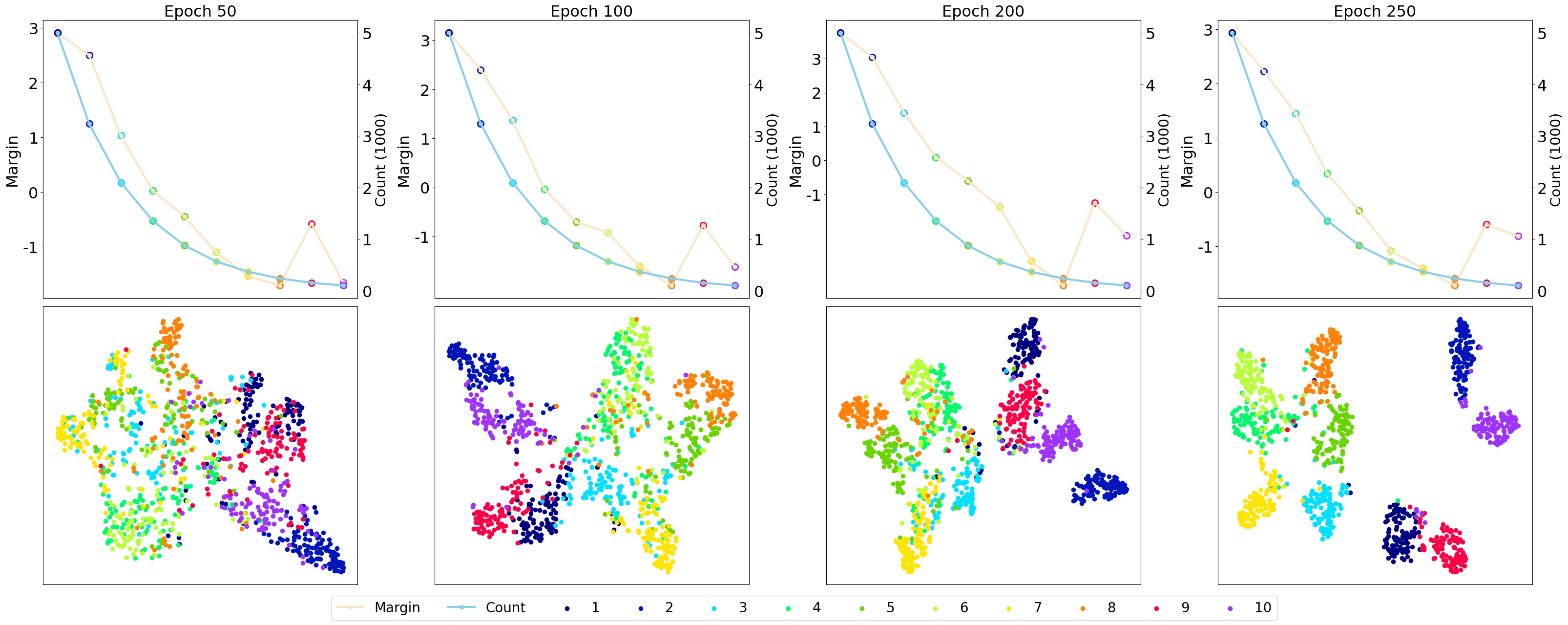}
    \caption{{Visualization of learned classification margins and feature distributions of meta set varied with training epochs.}}
    \label{fig:vis:margin_epoch}
\end{figure*}

\noindent\textbf{Learnable Weight Varied with Imbalance Ratios.}
{Figure~\ref{fig:vis:imbalance} depicts the learned margins generated by the margin generator under different imbalance ratios. We can observe that as the class index increases, corresponding to decreasing sample size, the generated margin consistently decreases, irrespective of the varying imbalance ratios.
Additionally, the variation in learned margins across different classes tends to increase as the imbalance ratio becomes more severe.
Both quantitative and qualitative analyses provide evidence that the margin generator effectively respects and adapts to various class distributions by automatically learning to assign appropriate margins.}

\noindent\textbf{Learnable Weight Varied in Training.}
{In Figure~\ref{fig:vis:margin_epoch}, the variations of classification margins and feature distributions of meta set are depicted across different training epochs.
It can be inferred that the margin generator continually adapts the classification margins based on the feature distributions of meta set during the training process.
For instance, in the case of class $10$, a relatively small classification margin is assigned to it since it is challenging to recognize at epoch 50.
However, as it becomes easier to recognize at epoch 250, the margin generator assigns a larger classification margin to it.
This observation supports the claim that the margin generator can dynamically regulate the classification margin based on the classification difficulty.}

\subsection{Behavior of hierarchical sampling.}
\begin{figure}[t]
 \centering
 \includegraphics[width=0.9\linewidth]{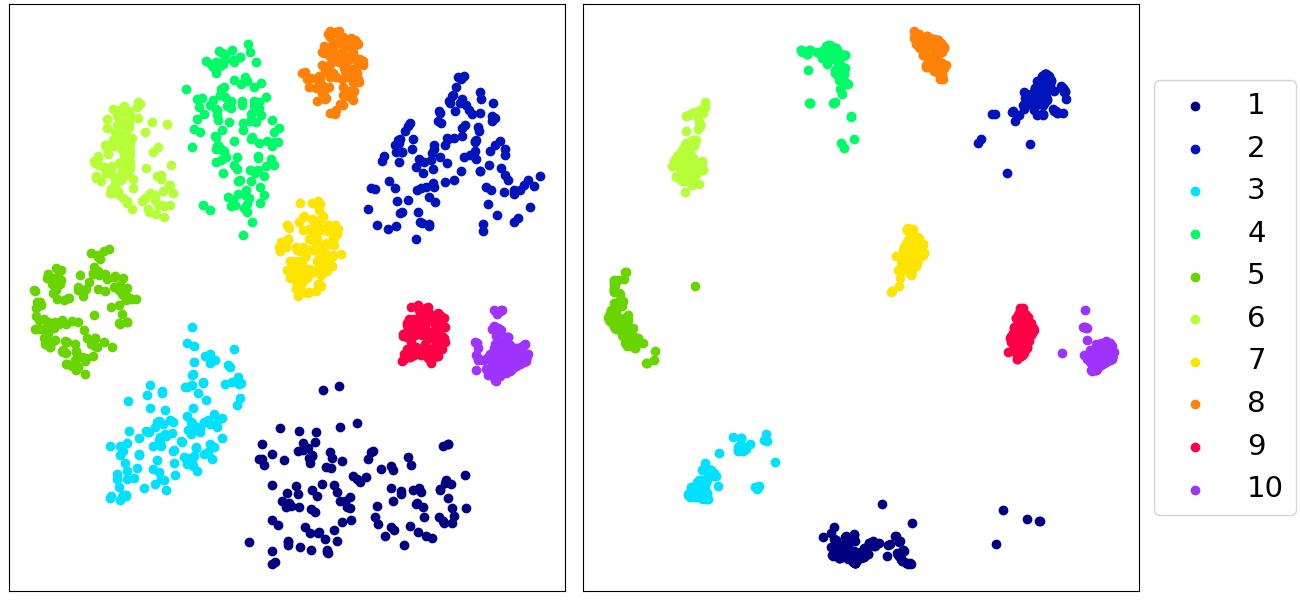}
    \caption{{The feature distribution of samples in the meta set, constructed using hierarchical sampling (left) and naive sampling (right) on CIFAR-10-N-LT with $\rho$=10 and $\lambda$=0.2.}}
    \label{fig:vis:tsne}
  \end{figure}

{We examine the efficacy of our hierarchical sampling strategy in enhancing the construction of meta sets. Figure~\ref{fig:vis:tsne} visually presents the feature distribution of the meta set generated using our hierarchical sampling method (left) and the meta set generated using a naive sampling method (right). The feature distribution of samples selected using hierarchical sampling demonstrates greater dispersion within distinct clusters compared to naive sampling. This indicates that the creation of a primary set prior to random sampling enables a more diverse meta data selection, encompassing both easy and challenging samples. As a result, this approach helps to mitigate biased learning on easier samples.}

\begin{table*}[t] 
  \caption{{Ablation studies conducted on CIFAR-10-N-LT, considering varying noise rates (0.1-0.5) and imbalance ratios (10, 100). The average accuracy is reported to evaluate model performance. BS represents Balanced Softmax.}}
  \captionsetup{position=bottom}
  \subfloat[Effect of key components.
  \label{tab:lc_mg}]{
  \setlength{\tabcolsep}{9.3pt}
  \renewcommand\arraystretch{1.1}
 \begin{tabular}{ccc|cc}
    \toprule
    $\mathcal{G}_{\theta_l}$ & $\mathcal{G}_{\theta_m}$ & BS    & Last  & Best \\
    \midrule
    \xmark & \xmark & \xmark & 52.73  & 60.43  \\
    \xmark & \cmark & \xmark & 71.51  & 76.84  \\
    \cmark & \xmark &  \xmark & 71.08  & 77.11  \\
    \cmark & \xmark & \cmark & 72.64  & 74.96  \\
    \cmark & \cmark &  \xmark & \textbf{79.73 } & \textbf{80.55 } \\
    \bottomrule
    \end{tabular}%
  }\hfill
  \subfloat[Ablations on sampling strategy.\label{tab:sampling}]{
  \setlength{\tabcolsep}{12.2pt}
  \renewcommand\arraystretch{1.1}
 \begin{tabular}{c|cc}
    \toprule
    Sampling & Last  & Best \\
    \midrule
    Naïve & 78.53 & 79.58  \\
    Hierarchical & \textbf{79.73} & \textbf{80.55} \\
    \bottomrule
    \multicolumn{3}{c}{} \\
    \multicolumn{3}{c}{} \\
    \end{tabular}%
  } \hfill
  \subfloat[Ablation on utilization of inputs.\label{tab:imp_exp}]{
  \setlength{\tabcolsep}{10.3pt}
  \renewcommand\arraystretch{1.1}
  \begin{tabular}{c|cc}
    \toprule
    Methods & Last  & Best \\
    \midrule
   Cross Entropy & 52.73  & 60.43  \\
    Implicit & 57.21  & 74.42  \\
    Explicit &  \textbf{79.73}  &  \textbf{80.55}  \\
    \bottomrule
    \multicolumn{3}{c}{}  \\
    \multicolumn{3}{c}{}  \\
    \end{tabular}%
    \vspace{2mm}
    } \\
  \subfloat[Ablation on optimization approach.\label{tab:sample-group}]{
  \setlength{\tabcolsep}{5.0pt}
  \renewcommand\arraystretch{1.1}
  \begin{tabular}{cc|cc}
    \toprule
    $\mathcal{G}_{\theta_l}$ & $\mathcal{G}_{\theta_m}$ & Last  & Best \\
    \midrule
    -     & -     & 52.73  & 60.43  \\
    Sample-wise  & Sample-wise    &  44.63 &  74.42 \\
    Sample-wise  & Group-wise    &  40.76 & 68.08  \\
    Group-wise   & Group-wise    & \textbf{79.73 } & \textbf{80.55 } \\
    \bottomrule
           \multicolumn{4}{c}{}  \\
    \end{tabular}%
  }\hfill
  \subfloat[Ablation on noisy label identifier.\label{tab:uncer}]{
  \setlength{\tabcolsep}{5.0pt}
  \renewcommand\arraystretch{1.1}
  \begin{tabular}{cc|cc}
    \toprule
    Loss rank  & Uncertainty & Last & Best\\
    \midrule
    \xmark     & \xmark    &  52.73  & 60.43 \\
    \xmark     & \cmark     & 72.91  & 77.18 \\
    \cmark     & \xmark     & 79.73 & 80.55 \\
    \cmark     & \cmark     & \textbf{80.62} & \textbf{81.23} \\
    \bottomrule
    \end{tabular}%
  }\hfill
  \subfloat[Ablations on meta net architecture.\label{tab:meta_archi}]{
  \setlength{\tabcolsep}{10.0pt}
  \renewcommand\arraystretch{1.1}
  \begin{tabular}{c|cc}
    \toprule
    Architecture & Last  & Best \\
    \midrule
    Vector & 79.50  & 80.35  \\
    3 layer MLPs & \textbf{79.73 } & \textbf{80.55 } \\
    \bottomrule
        \multicolumn{3}{c}{} \\
         \multicolumn{3}{c}{} \\
    \end{tabular}%
    }
\label{tab:ablations} 
 \vspace{-3mm}
\end{table*}


\section{Ablation Studies.}

\noindent\textbf{Effect of Label Corrector.}
{We construct a model variant wherein the label corrector component is removed to assess its efficacy. Table~\ref{tab:lc_mg} illustrates that the average final accuracy of this variant decreases by $8.22\%$ compared to the complete dynamic loss configuration. This notable decrease in performance provides compelling evidence of the effectiveness of the label corrector.}

\noindent\textbf{Effect of Margin Generator.}
{As indicated in Table~\ref{tab:lc_mg}, to assess the efficacy of the margin generator, we initially utilize only the label corrector, which yields an average final accuracy of $71.08\%$. Subsequently, we incorporate Balanced-Softmax to address class imbalance, which only results in a marginal improvement of $1.56\%$ in accuracy. Finally, by implementing our margin generator, we are able to further improve the accuracy to $79.73\%$. These results serve as evidence supporting the importance of utilizing a dynamic margin in managing long-tailed noisy data.}

\noindent\textbf{Effect of Hierarchical Sampling.}
{Table~\ref{tab:sampling} demonstrates that the substitution of hierarchical sampling with naive random sampling leads to a reduction of up to $1.20\%$ in average final accuracy. This observation suggests that the meta set created using hierarchical sampling possesses a more comparable distribution to the test set.}

\noindent\textbf{Effect of Class-specific Label Corrector.}
{To validate the class-specific design of our label corrector, we constructed a class-agnostic variant and evaluated its performance on CIFAR-10-N, a dataset containing $40\%$ asymmetric noise with varying noise rates across different classes. Our results demonstrate that the class-specific label corrector significantly outperforms its class-agnostic counterpart by a large margin of $3.95\%$ in terms of accuracy ($94.51\%$ vs. $90.56\%$), thus providing clear evidence for the effectiveness of our class-specific design.}

\noindent\textbf{Effect of Explicit Utilizing Given and Predicted labels.} 
{To assess the impact of explicitly utilizing known information, particularly given labels and predicted labels, we conducted a comparative analysis between the implicit and explicit approaches in utilizing such information. The results, as presented in Table~\ref{tab:imp_exp}, demonstrate that the explicit utilization of known information resulted in a remarkable 22.52\% increase in final accuracy and a substantial 6.13\% improvement in best accuracy compared to the implicit approach. These findings indicate that simplifying the task of the meta net through the explicit utilization of known information can significantly enhance the performance of the classifier. Furthermore, the notable discrepancy between the final accuracy and the best accuracy underscores the contribution of explicit utilization of known information in facilitating the convergence of the meta net.}

\noindent\textbf{Effect of Group-wise Optimization.} 
{To validate the efficacy of the group optimization approach employed in our method, we conducted experiments comparing it with sample-wise optimization. The results, presented in Table~\ref{tab:sample-group}, reveal that utilizing a sample-wise approach with both the label corrector and margin generator hindered classifier convergence, leading to significantly lower final accuracy compared to the best achieved accuracy. Similarly, employing the margin generator in a group-wise manner alone failed to yield convergence. However, when extending the group-wise approach to the label corrector, a final accuracy of 79.73\% was attained. These findings highlight the significance of optimizing the meta net task, as it enhances learning efficiency and improves the classifier's robustness.}

\begin{table}[t]
  \centering
   \setlength{\tabcolsep}{2.5pt}
  \renewcommand\arraystretch{1.2}
  \caption{{Accuracy (\%) on CIFAR10-N-LT with varying imbalance ratio. WRN-28-10 used as the classifier.}}
   \begin{tabular}{l|ccc|ccc|c}
    \toprule
    \multicolumn{1}{c|}{\textbf{Noise Rate}} & \multicolumn{3}{c|}{\textbf{0.2}} & \multicolumn{3}{c|}{\textbf{0.4}} & \multirow{2}[4]{*}{\textbf{Avg.}} \\
\cmidrule{1-7}    \multicolumn{1}{c|}{\textbf{Imbalance Ratio}} & \textbf{10}    & \textbf{50}    & \textbf{200}   & \textbf{10}    & \textbf{50}    & \textbf{200}   &  \\
    \midrule
    Cross Entropy    & 78.03  & 65.53  & 42.06  & 63.04  & 47.56  & 29.09  & 54.22  \\
    HAR-DRW~\cite{cao2020heteroskedastic} & 88.81  & 82.74  & 73.98  & 84.03  & 75.36  & 63.95  & 78.15  \\
    FSR~\cite{zhang2021learning}   & 85.70  & 77.40  & 65.50  & 81.60  & 69.80  & 49.50  & 71.58  \\
    INOLML~\cite{hoang2022maximising} & 90.10  & 80.10  & 66.60  & 89.10  & 78.10  & 61.60  & 77.60  \\
    \midrule
    \textbf{Dynamic Loss}  & \textbf{92.40 } & \textbf{83.56 } & \textbf{76.72 } & \textbf{90.46 } & \textbf{78.54 } & \textbf{66.47 } & \textbf{81.36 } \\
    \bottomrule
    \end{tabular}%
  \label{tab:wrn-28-10}%
\end{table}%

\noindent\textbf{Test with MC-Dropout Uncertainty.} 
{To further evaluate the generalizability of our proposed approach, we conducted an assessment by substituting the loss rank metric with MC-Dropout uncertainty, renowned for its simplicity and effectiveness in uncertainty estimation. 
The results of this evaluation are presented in Table~\ref{tab:uncer}.
Initially, we exclusively fed the MC-Dropout uncertainty to the meta net, which yielded a final accuracy of 72.91\% and a best accuracy of 77.18\%. However, this performance was evidently inferior compared to when the loss rank was provided to the MetaNet. Subsequently, we incorporated both the loss rank and uncertainty as inputs to the meta net, resulting in the highest performance. Specifically, this configuration achieved a final accuracy of 80.62\% and a best accuracy of 81.23\%. These findings suggest that uncertainty estimation can serve as a significant supplement to loss rank in the context of learning with noisy labels.}

\noindent\textbf{Effect of Meta Net Architecture.}
{In order to validate the architecture design of the meta net, we simplified the label corrector and margin generator by replacing them with a $R$-length and $C$-length learnable vector, respectively. As shown in Table~\ref{tab:meta_archi}, this modification resulted in a noticeable performance drop of $0.23\%$. Our experimental observation suggests that this is due to the fact that MLPs are able to quickly learn appropriate label weights and per-class margins, whereas the learnable vectors suffer from slow convergence.}

\noindent\textbf{Test on More Classifiers.}
{To demonstrate the broad applicability of our method, we conducted additional evaluations using the Wide-ResNet-28-10 (WRN-28-10) architecture~\cite{DBLP:conf/bmvc/ZagoruykoK16}. Specifically, we set the imbalance ratios to 10, 50, and 200, and the noise rates to 0.2 and 0.4. The mean accuracy results are presented in Table~\ref{tab:wrn-28-10}. Our findings indicate that our method surpasses the performance of HAR, FSR~\cite{zhang2021learning}, and INOLML~\cite{hoang2022maximising} on both CIFAR-10-N-LT datasets. These results provide compelling evidence supporting the effectiveness of our dynamic loss across different classifier architectures.}

\noindent\textbf{Training Time Analysis}
{As training time is a critical concern in meta learning, we evaluated the total training time of our methods following the approach taken in DivideMix. Thanks to the use of FaMUS~\cite{xu2021faster} and CurveNet~\cite{jiang2021delving} to accelerate the training speed of meta learning, we were able to train a model in approximately 7.2 hours using an NVIDIA GTX 1080 Ti. This is slightly slower than DivideMix with Nvidia V100 GPU (5.2 hours), but it provides evidence of the efficiency of our method.}

\section{Conclusions}
{This work introduces a novel dynamic loss for robust learning from long-tailed data with noisy labels. The dynamic loss consists of a learnable label corrector and margin generator, which jointly correct noisy labels and adjust classification margins to guide classifier learning. The meta net and classifier are co-optimized through meta-learning using a hierarchical sampling strategy that provides unbiased yet diverse meta data. Extensive evaluations on both synthetic and real-world data demonstrate the effectiveness of our dynamic loss, which exhibits high adaptability and robustness to various types of data biases.}

 \bibliographystyle{IEEEtran}
\bibliography{ref.bib}

\end{document}